\documentclass[accepted]{uai2026} % after acceptance, for a revised version; 
\usepackage[utf8]{inputenc} % allow utf-8 input
\usepackage[T1]{fontenc}    % use 8-bit T1 fonts
\usepackage{hyperref}       % hyperlinks
\usepackage{url}            % simple URL typesetting
\usepackage{booktabs}       % professional-quality tables
\usepackage{amsfonts}       % blackboard math symbols
\usepackage{nicefrac}       % compact symbols for 1/2, etc.
\usepackage{microtype}      % microtypography
\usepackage{xcolor}         % colors

\usepackage{array}
\usepackage{longtable}
\usepackage{amssymb}
\usepackage{bm}
\usepackage{tcolorbox}
\usepackage{pifont}
\usepackage[labelformat=simple]{subcaption}
\usepackage{amsthm}
\usepackage{thmtools}
\usepackage{wrapfig}
\usepackage{mathtools}
\usepackage{enumitem}
\usepackage{wrapfig}

\renewcommand\thmcontinues[1]{Continued}

\newcommand{\vtheta}{\bm{\theta}}
\newcommand{\vvtheta}{\bm{\vartheta}}
\newcommand{\given}{\, | \,}

\newcommand{\bE}{\mathbb{E}}
\newcommand{\Ree}{\mathbb{R}}
\newcommand{\bV}{\mathbb{V}}

\newcommand{\tu}{\operatorname{TU}}
\newcommand{\au}{\operatorname{AU}}
\newcommand{\eu}{\operatorname{EU}}

\newcommand{\cP}{\mathcal{P}}

\newcommand{\cY}{\mathcal{Y}}
\newcommand{\cX}{\mathcal{X}}
\newcommand{\tr}{\mathrm{tr}}
\newcommand{\bx}{\boldsymbol{x}}
\newcommand{\by}{\boldsymbol{y}}
\newcommand{\bX}{\boldsymbol{X}}
\newcommand{\bY}{\boldsymbol{Y}}
\newcommand{\bc}{\boldsymbol{c}}

\newtheorem{proposition}{Proposition}[section]
\newtheorem{definition}{Definition}[section]

\usepackage[american]{babel}
\usepackage{natbib} % has a nice set of citation styles and commands
\usepackage{mathtools} % amsmath with fixes and additions
\usepackage{booktabs} % commands to create good-looking tables
\usepackage{tikz} % nice language for creating drawings and diagrams

\title{An Axiomatic Assessment of Entropy- and Variance-based Uncertainty Quantification in Regression}

\author[1,2]{Christopher Bülte}
\author[1,2]{Yusuf Sale}
\author[1,2]{Timo Löhr}
\author[1,2]{Paul Hofman}
\author[1,2,3,4]{Gitta Kutyniok}
\author[1,2,5]{Eyke Hüllermeier}

\affil[1]{%
LMU Munich
}
\affil[2]{%
Munich Center for Machine Learning (MCML)
}
\affil[3]{%
DLR-German Aerospace Center
}
\affil[4]{%
University of Troms\o
}
\affil[5]{%
German Research Center for Artificial Intelligence (DFKI, DSA) 
}

\begin{document}
\maketitle
\begin{abstract}
Uncertainty quantification is crucial in machine learning, yet most (axiomatic) studies of uncertainty measures focus on classification, leaving a gap in regression settings with limited formal justification and evaluations. 
In this work, we provide a formal way of representing uncertainty in continuous space, using a general parametric formulation, allowing for tractable analysis and evaluation of uncertainty measures. Within this framework, we propose a set of axioms that enable rigorous assessment of total, aleatoric, and epistemic uncertainty measures. Together, this allows for a theoretical examination of uncertainty measures and their corresponding properties.
As a specific example, we compare the widely used entropy- and variance-based measures with respect to established predictive models and analyze their limitations and challenges in uncertainty quantification. Our work provides a principled way to understand and develop uncertainty measures in supervised regression, offering theoretical insights and practical guidelines for reliable uncertainty assessment.
\end{abstract}

\section{Introduction}\label{sec:intro}
Uncertainty quantification (UQ) has emerged as an active field in contemporary machine learning research and practice\textemdash{}particularly in supervised regression settings. Mostly due to the application of AI systems to safety-critical tasks, research has focused on quantifying uncertainty in domains such as weather forecasting \citep{NeuralNetworksforPostprocessingEnsembleWeatherForecasts, bülte2024uncertaintyquantificationdatadrivenweather}, energy systems \citep{en16083522}, autonomous driving \citep{michelmore2018evaluatinguncertaintyquantificationendtoend} or healthcare \citep{uq_mri,lohr2024towards}. Having a thorough understanding of uncertainty is key in these domains.

In the aforementioned supervised learning tasks, the focus is usually on \emph{predictive uncertainty}, which describes the uncertainty of the outcome $\by \in \mathcal{Y}$ given underlying data $\bx \in \mathcal{X}$. There has been a particular emphasis on distinguishing \emph{aleatoric} and \emph{epistemic} uncertainty, as well as the decomposition of \emph{total} uncertainty into its aleatoric and epistemic components \citep{hullermeier2021aleatoric}. Aleatoric uncertainty captures the inherent randomness in the data-generating process. As it represents variability that cannot be reduced even with more data, it is often referred to as \emph{irreducible} uncertainty. Sources of aleatoric uncertainty include inherent randomness in physical systems or measurement errors. In contrast, epistemic uncertainty arises from a lack of knowledge or understanding about the underlying data-generating process, which can be reduced by acquiring additional data or improving the model itself. Therefore, epistemic uncertainty is also referred to as \emph{reducible} uncertainty.

When analyzing uncertainty, it is of particular importance to distinguish between the \emph{representation} and the \emph{quantification} of uncertainty. The latter refers to the task of measuring (types of) uncertainty conveyed by a given representation. While aleatoric uncertainty can effectively be represented by probability distributions, representing epistemic uncertainty poses significant challenges. In fact, there is a general consensus that epistemic uncertainty \emph{cannot} be represented by means of classical probability theory. Consequently, higher-order formalisms are adopted, such as second-order distributions (distributions of distributions) or credal sets (sets of probability distributions) \citep{levi1980enterprise, walley1991statistical}.

Various approaches have been proposed in the literature, addressing the inherent difficulty of \emph{evaluating} the quality of uncertainty measures \citep{hullermeier2021aleatoric, 10.1007/978-3-030-50146-4_41}. One such prominent approach is the \emph{axiomatic} evaluation of uncertainty quantification, which assesses the suitability of measures for total, aleatoric, and epistemic uncertainty based on a predefined set of axioms. However, evaluations primarily focus on the supervised classification setting \citep{hullermeier2022quantification, wimmer2023quantifying, sale2024label, sale2023second, sale2023volume}, although many machine learning problems are actually within the regression framework, from classical regression to inverse problems. In contrast to classification, the underlying outcome space of supervised regression is generally unbounded, preventing a direct transfer of results. Although some research has analyzed uncertainty quantification for the regression setting \citep{berry2024efficientepistemicuncertaintyestimation, malinin2020regressionpriornetworks, amini2020deep}, there is a noticeable lack of \emph{formal} justification for many proposed approaches, as well as an absence of evaluations grounded in an axiomatic framework.

\textbf{Contributions.} In this paper, we propose a formal uncertainty representation framework that allows for tractable analysis, along with a set of formalized axioms that provide a way to evaluate uncertainty measures theoretically.
To represent uncertainty, we propose a general parametric family that admits a natural characterization of epistemic uncertainty and contains many widely used predictive machine learning methods. 
We then provide a set of six axioms that uncertainty measures should satisfy, specifically designed and adapted to accommodate the continuous regression setting. Finally, we analyze the most commonly used uncertainty measures, based on entropy or variance, with respect to the proposed axioms, highlighting differences and identifying pitfalls. Our analysis reveals that both entropy- and variance-based measures, despite their widespread adoption, exhibit fundamental shortcomings: each violates specific axioms in ways that directly translate to undesirable and potentially misleading behavior regarding uncertainty quantification.
Our framework provides a principled foundation for both evaluating existing uncertainty measures and guiding the design of new ones, directly addressing a gap that has left practitioners without formal criteria for this choice.

\section{Representing and quantifying uncertainty in regression}
\label{sec:quant}
In the following, we denote by $\mathcal{X} \subseteq \mathbb{R}^{k}$, $\mathcal{Y} \subseteq \mathbb{R}^{d}$, and $\Theta \subseteq \Ree^p$ the (real-valued) feature, target, and parameter space, respectively. Furthermore, define the probability measures $\mathcal{P}(\mathcal{X}),\mathcal{P}(\mathcal{Y}),\mathcal{P}(\Theta)$ over the corresponding measurable spaces.
Assume we have training data $\mathcal{D} = \{\boldsymbol{x}_i, \boldsymbol{y}_i  \}_{i=1}^n \in (\mathcal{X} \times \mathcal{Y})^n$, where for $i \in \{1, \dots, n\}$, each pair $(\bx_i,  \by_i)$ is a realization of the i.i.d. random variables $(\bX_i, \bY_i)$, distributed according to some probability measure $P$. Consequently, a given feature vector $\bm{x} \in \mathcal{X}$ induces a conditional probability distribution $P(\cdot \given \bm{x}) \in \mathcal{P}(\mathcal{Y})$.

\subsection{Uncertainty Representation}
\label{subsec:ur}
In general, aleatoric uncertainty can naturally be represented by the induced conditional probability distribution ${P(\cdot \mid \bx)}$. Epistemic uncertainty, on the other hand, requires a higher-order representation, for which we adopt a second-order distribution \citep{hullermeier2021aleatoric}. In the classification setting, this definition is straightforward, as one is working with predictions in the $K-1$-dimensional probability simplex $\Delta_K$, on top of which the second-order distribution $Q$ is placed, i.e., $Q \in \cP(\Delta_K)$.

However, in the regression setting, as the prediction space $\mathcal{Y}$ is continuous and unbounded, the corresponding definitions turn out to be more complex.
While one can formally define the second-order distribution as an abstract distribution over probability measures, i.e., $Q \in \cP(\cP(\cY))$, this definition cannot be practically used to evaluate uncertainty measures.
To make uncertainty quantification and uncertainty evaluation feasible, we propose to use parametric first-order distributions and place a second-order distribution over the corresponding parameter space $\Theta$, allowing for operationalization of the second-order distribution.

In particular, we consider a parametric family of predictive distributions
\begin{equation}
    \label{eq:parametric_family}
    \{P(\cdot \mid \vtheta) : \vtheta \in \Theta\} \subseteq \mathcal{P}(\mathcal{Y}),
\end{equation}
where the distributional family itself is fixed and all uncertainty resides in the parameter vector $\vtheta \in \Theta$. The dependence on the covariates is then captured by a covariate-dependent second-order distribution
\begin{equation}
    \label{eq:second_order}
    Q(\bx) \in \cP(\Theta),
\end{equation}
representing the (epistemic) state of knowledge about the first-order parameters at a given instance $\bx \in \cX$.

In this setting, if $Q(\bx) = \delta_{\vtheta(\bx)}$, we recover the parametric predictive distribution $P(\cdot \mid \bx) = P(\cdot \mid \vtheta(\bx))$ with covariate-dependent parameter vector $\vtheta(\bx) \in \Theta$. Here, the dependence on the covariates is typically modeled via $\theta^k(\bx) = h_k(\eta_k(\bx)),\ k = 1,\ldots, p$, where $\eta_k(\bx)$ are (parameter-specific) regression predictors and the response function $h_k: \Ree \to \Ree$ maps the predictors to the appropriate parameter space, for example with the exponential function for nonnegative parameters. This case includes generalized linear and generalized additive models \citep{kneib2023rage}, but also neural network-based probabilistic methods, when $\eta_k(\bx) = \mathcal{F}^k_\phi(\bx)$ is parameterized with a neural network $\mathcal{F}_\phi^k$ with weights (and biases) $\phi$. In this setting, each distributional parameter $\theta^k$ is a (possibly nonlinear) function of the covariates $\bx$, allowing for huge flexibility regarding the types and complexity of the target distribution, as the parameters can represent any distributional aspects, such as location, scale, or shape \citep{kneib2023rage}. A non-degenerate $Q(\bx)$, in contrast, additionally expresses epistemic uncertainty about the first-order parameters themselves, including conjugate prior settings such as in deep evidential regression \citep{amini2020deep}.

More formally, let ${\varphi:\Theta\to\mathcal P(\mathcal Y): \, \varphi(\vtheta)=P_{\vtheta} \coloneq P(\cdot \mid \vtheta)}$ be a measurable map, which essentially transforms the parameter vector $\vtheta$ into a (specific) probability measure $P_{\vtheta}$. Given some $\vtheta \in \Theta$, aleatoric uncertainty is naturally available via $P_{\vtheta}$. Epistemic uncertainty, by contrast, requires a higher-order formalism, which we represent through the probability distribution $Q(\bx) \in \mathcal{P}(\Theta)$ over $\Theta$.
By defining $Q^\varphi(\bx) \coloneq \varphi_{\#}Q(\bx) \in \cP(\cP(\cY))$ as the push-forward of $Q(\bx)$ under $\varphi$ (the law of the random first-order distributions $P_{\vvtheta}$, when $\vvtheta \sim Q(\bx)$), we then implicitly obtain the (second-order) probability distribution over the space $\cP(\cY)$.
 
Finally, we can define the predictive (mixture) distribution, as
\begin{equation}
\overline{P}(\bx) \;=\; \int_\Theta P_{\vtheta} \, Q(\bx)(d\vtheta)\;\in\;\mathcal{P}(\mathcal{Y}).
\end{equation}
Here, we assume that all first-order measures are dominated by the Lebesgue measure $\mu$, i.e., for all $\vtheta \in \Theta$ we have $P_{\vtheta} \ll \mu$ (and hence $\overline{P}(\bx) \ll \mu$), and obtain the corresponding densities, denoted as $p_{\vtheta}$ and $\overline{p}(\cdot \mid \bx)$, respectively.
% The same holds for the distributions over the space $\Theta$.
As all subsequent definitions, axioms, and results are understood to hold pointwise for a given instance $\bx \in \cX$, we henceforth drop the dependence on the covariates $\bx$ for ease of notation.
 
Note that in this setting, we treat the distributional family as fixed and assume that all uncertainty resides in the parameters $\vtheta$. While this specification might initially seem restrictive, deriving theoretical statements about uncertainty measures requires a trade-off between tractability and generalization. Still, our setting allows for great flexibility regarding the distributional family (e.g. multivariate-, nonnegative- or skewed distributions) and has been utilized for regression tasks in various applications, such as geodesy \citep{KIANISHAHVANDI2025105818}, forecasting of weather extremes \citep{schulzMachineLearningMethods2022, friederichs2012forecast}, photometric redshift estimation \citep{refId0} or photovoltaic power forecasting \citep{mayer2025postprocessingensemblephotovoltaicpower, GNEITING202372}.
Furthermore, our setup also includes many commonly used machine learning approaches for uncertainty representation, including (deep) ensembles, distributional regression, or deep evidential regression.

\subsection{Uncertainty Quantification}
\label{subsec:uq}
\begin{figure}[ht]
\begin{minipage}[c]{0.01\textwidth}
    \phantom{A} \\
    \vspace{0.1em}
    \rotatebox{90}{Aleatoric} \\
    \vspace{2.5em}
    \rotatebox{90}{Epistemic}
\end{minipage}
\hfill
\begin{minipage}[c]{0.23\textwidth}
    \centering
    \phantom{p}Variance-based \\
    \vspace{0.2em}
    \includegraphics[height=7.2em]{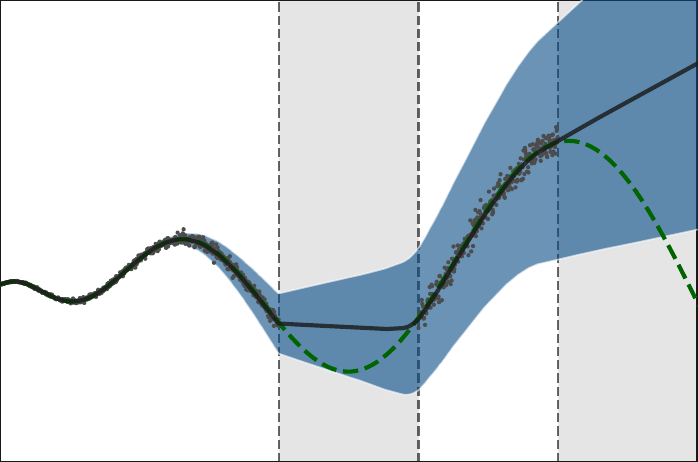} \\
    \vspace{0.5em}
    \includegraphics[height=7.2em]{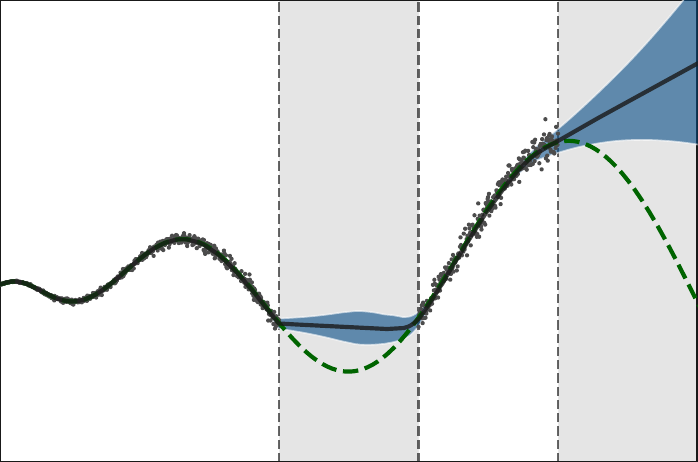}
\end{minipage}
\begin{minipage}[c]{0.23\textwidth}
    \centering
    Entropy-based \\
    \vspace{0.2em}
    \includegraphics[height=7.2em]{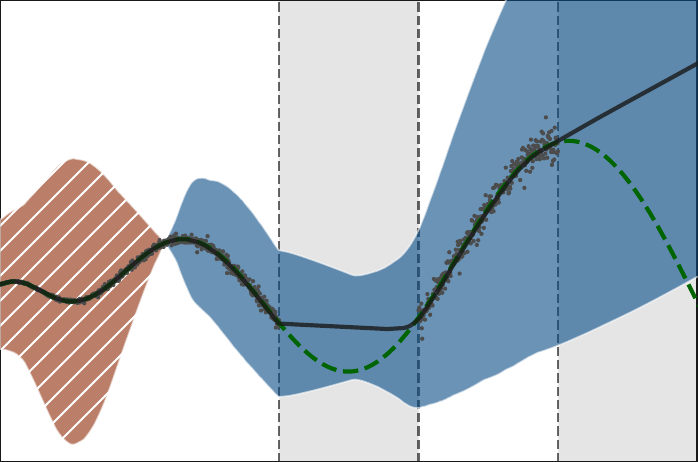} \\
    \vspace{0.5em}
    \includegraphics[height=7.2em]{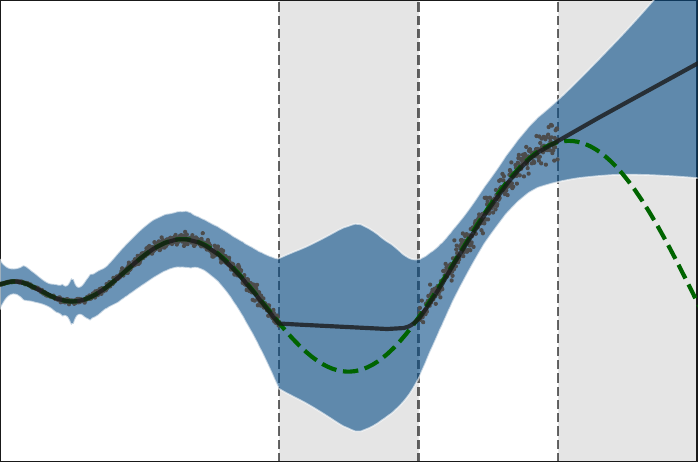}
\end{minipage}

\hspace{0.61em}
\begin{minipage}[r]{0.1\textwidth}
    \includegraphics[height=2.41em]{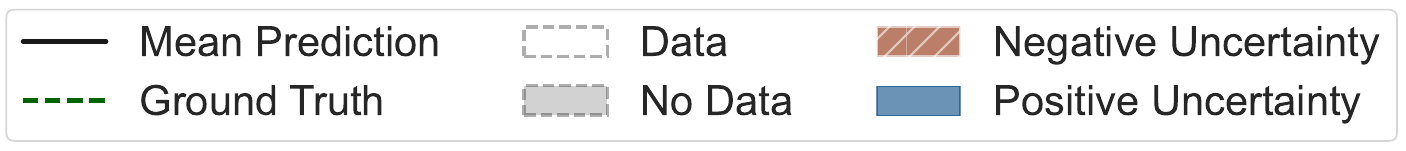}
\end{minipage}
    \caption{Comparison of variance and entropy-based uncertainty measures for deep ensembles on a synthetic example. Note in particular the occurrence of negative uncertainty values for the entropy-based measure.}
    \label{fig:ensemble_comparison}
        \vspace{-0.3cm}
\end{figure}

In this section, we recall the two predominant sets of measures used in the literature to \emph{quantify} total (TU), aleatoric (AU), and epistemic uncertainty (EU), namely entropy-based and variance-based measures.
In the following, consider a first-order distribution $P_{\vvtheta} \in \cP(\cY)$ with a random parameter vector $\vvtheta \sim Q$ and assume that $Q$ satisfies the necessary integrability conditions ensuring the existence of all uncertainty measures defined below.

\emph{Entropy-based measures. } The most widely adopted measures for quantifying total, aleatoric, and epistemic uncertainty associated with a second-order distribution in the classification setting rely on Shannon entropy and its decomposition into conditional entropy and mutual information \citep{houlsby2011bayesian, kendall2017uncertainties, charpentier2022disentangling}. This decomposition can be directly transferred to the regression setting by considering differential entropy \citep{malinin2020regressionpriornetworks,postels2021hiddenuncertaintyneuralnetworks}, which however has structurally different properties.

Let $\bY$ denote the outcome with marginal density $p(\by) = \int_{\Theta} p(\by \mid \vtheta) \, dQ(\vtheta)$. Then, its differential entropy is given by $H(\bY) = - \int_\cY p(\by) \log p(\by) \, d\by$. 
Following \cite{houlsby2011bayesian}, we can utilize conditional differential entropy to establish the following well-known decomposition:
\begin{equation}    \label{eq:entropy_uncertainty}
    \underbrace{H(\bY)}_{\text{Entropy}} = \underbrace{H(\bY \mid P_{\vvtheta})}_{\text{conditional entropy}} + \underbrace{I(\bY, P_{\vvtheta}).}_{\text{mutual information}}
\end{equation}

Then, the uncertainty measures can be defined as
\begin{align}
    \tu(Q) &= H(\overline{P}), \\ \au(Q) &=  \bE_Q[H(\bY \mid P_{\vvtheta})], \\
    \eu(Q) &= \mathbb{E}_Q[D_\text{KL}(P_{\vvtheta} \|\overline{P})],   
\end{align}
where $D_\text{KL}(\cdot\| \cdot)$ denotes the Kullback-Leibler (KL) divergence and $\overline{P}$ is the previously introduced predictive mixture distribution. 

Intuitively, AU can be assessed by fixing a first-order distribution $P_{\vtheta}$, thus removing all epistemic uncertainty. However, as $\vtheta$ is not exactly known, we take the expectation over the second-order distribution $Q$. Epistemic uncertainty, defined as the mutual information, quantifies the potential reduction in uncertainty of $\bY$ through observing $P_{\vvtheta}$ \citep{ash1990information}.
The entropy-based measures require absolutely continuous distributions with respect to the Lebesgue measure, i.e., $P_{\vtheta} \ll \mu, \ \forall \vtheta \in \Theta$, which we assume throughout.

\emph{Variance-based measures.} 
In classification, another set of commonly used uncertainty measures is based on the law of total variance \citep{sale2023second}. 
Even more naturally, this decomposition extends to the real-valued case and is widely applied in regression settings \citep{amini2020deep, valdenegrotoro2022deeperlookaleatoricepistemic, laves2021recalibrationaleatoricepistemicregression}.

To accommodate for the (possibly) multivariate setting, we use the scalar generalization of the variance via the trace of the covariance matrix, i.e., $\bV(\bY) \coloneq \bE \left[\|\bY - \bE[\bY]\|_2^2 \right] = \tr(\mathrm{Cov}(\bY))$. From this, we can leverage the law of total variance to decompose total uncertainty into its aleatoric and epistemic components:
\begin{equation}\label{eq:variance}
    \underbrace{\bV(\bY)}_{\text{total}} = \underbrace{\bE_Q[\bV(\bY \given \vvtheta)]}_{\text{aleatoric}} + \underbrace{\bV_Q(\bE[\bY \given \vvtheta])}_{\text{epistemic}}.
\end{equation}
In the multivariate case, the uncertainty measures correspond to the sum of all marginal uncertainties; in the univariate case, this corresponds to the commonly-used variance-based decomposition \citep{amini2020deep}.

Similar to the entropy-based measures, AU is the variance in the outcome variable when removing epistemic uncertainty by fixing a (known) parameter $\vvtheta$. 
However, as $\vvtheta$ is unknown, we take the expectation with respect to $Q$. Epistemic uncertainty, on the other hand, is given by the variance of the mean prediction, e.g., the uncertainty induced by the second-order distribution $Q$.
The variance-based measures require finite second moments, i.e., $\bE[\|\bY\|_2^2] < \infty$ to exist, which, in contrast to the classification setting, is not necessarily fulfilled, for example, for heavy-tailed distributions.

While both types of measures provide meaningful estimates of the different kinds of uncertainty and fulfill an additive decomposition, i.e., $\tu = \au + \eu$, they differ significantly in their definition, assumptions, and properties, as visualized in \autoref{fig:ensemble_comparison}.

\subsection{Uncertainty evaluation}
% %
Although many uncertainty measures have been proposed, it remains an open question how they can be properly justified or evaluated. Generally speaking, such a justification could be either of a theoretical or empirical nature.
The question of how to empirically evaluate (aleatoric and epistemic) uncertainty in practice is highly nontrivial, in particular since there is typically no directly observable ground truth for uncertainty.
Unlike standard predictive tasks, where predictions can be compared against actual observed outcomes, no such ground truth is available when evaluating uncertainty.
As a consequence, uncertainty evaluation typically relies on proxies or indirect validation strategies.
In particular, the focus is often on downstream tasks such as selective prediction, active learning, and out-of-distribution detection. 
While these evaluation protocols have become standard, they are not without criticism \citep{li2025position}.

Given the inherent difficulties of empirical evaluation, a complementary evaluation approach is of a theoretical nature, namely, assessing whether a given uncertainty measure satisfies qualitative properties one would expect from its intended semantics. 
The importance of this (axiomatic) perspective is highlighted by recent results indicating that prominent uncertainty measures do not always satisfy even basic desiderata \citep{wimmer2023quantifying}.
In this spirit, our paper is centered around an axiomatic evaluation approach.

\section{Formal Properties for Uncertainty Measures}
Evaluating measures based on a set of axioms has been standard practice in the (uncertainty) literature \citep{pal1993uncertainty, bronevich2008axioms} and has been adopted by the machine learning community \citep{hullermeier2022quantification}.

In the context of supervised classification, recent works have proposed axiomatic frameworks for second-order uncertainty \citep{wimmer2023quantifying, sale2023second} with subsequent contributions extending these foundations \citep{sale2023volume} or proposing novel types of uncertainty measures \citep{kotelevskii2025from, hofman2024quantifying}.
Moving beyond classification, we propose a set of axioms specifically tailored to the regression setting and the continuous structure of the target space $\mathcal{Y}$.
The transfer of existing axioms (A0--A3) is inherently asymmetric: classification admits a natural notion of maximal epistemic uncertainty through the uniform distribution, which has no counterpart in regression, requiring a mathematical reformulation. Our proposed novel axioms regarding translation invariance (A4) and scale-order equivariance (A5) further exploit the metric and algebraic structure of $\Ree^d$, which is absent on a discrete label space.
Before introducing this set of axioms, we first establish the relevant notation and preliminaries.

\begin{definition}[Translation-invariance]
\label{def:translation}
Assume that the family $\{P_{\vtheta}: \vtheta \in \Theta\}$ is closed under translation, i.e., for every $\boldsymbol{c} \in \Ree^{d}$ there exists a parametrization $\tau_{\boldsymbol{c}}: \Theta \to \Theta, \ \vtheta \mapsto \vtheta_{\boldsymbol{c}}$ such that $p_{\tau_{\boldsymbol{c}}(\vtheta)}(\boldsymbol{y}) = p_{\vtheta}(\boldsymbol{y}-\boldsymbol{c}), \ \forall  \boldsymbol{y} \in \mathcal{Y}$. Define $Q_{\bc}$ as the pushforward of $Q \in \mathcal{P}(\Theta)$ under this mapping, i.e., $Q_{\bc} \coloneq \tau_{\bc\#}Q$. 
Then, an uncertainty measure $M$ is called \textbf{translation-invariant} if
\[
M(Q) = M(Q_{\bc}), \quad \forall Q \in \mathcal{P}(\Theta), \ \forall \bc \in \Ree^{d}.
\]
\end{definition}
This definition formalizes the common notion of translation-invariance to a second-order uncertainty measure and implies that the corresponding measure remains unchanged under a constant location shift in the first-order distribution.

\begin{definition}[Scale-order equivariance]
\label{def:scale_order}
    For $a>0$, define the scaling map $S_{a}$ implicitly via
    \[
    Y \sim P \implies a Y  \sim S_{a} P.
    \]
Assume that the predictive family $\{P_{\vtheta}: \vtheta \in \Theta\}$ is closed under scaling, i.e., there exists a measurable map ${\tau_{a} : \Theta \to \Theta}$ such that $P_{\tau_{a}(\vtheta)} = S_{a} P_{\vtheta}$. Define $Q^{a}$ as the pushforward of $Q \in \mathcal{P}(\Theta)$ under this mapping, i.e., $Q^{a} \coloneq \tau_{a \#}Q$.
Then, an uncertainty measure $M$ is called \textbf{scale-order equivariant} if 
\[
M(Q^{a}) = g(a) M(Q) + c(a), \quad a > 0,
\]
for functions $g:\Ree_{+}  \to \Ree_+, \ c: \Ree_+ \to \Ree$, with $g,c$ nondecreasing in $a$.
\end{definition}
Similarly, this definition formalizes the notion of scale-invariance. However, here we only require monotonicity, i.e., that the ordering of the uncertainty measure remains unchanged under a scaling of the first-order distribution.

Further, let $\delta_{\bm\theta} \in \mathcal{P}(\Theta)$ denote the Dirac measure at $\bm \theta \in \Theta$.
For $P_{\vtheta_u}, P_{\vtheta_l}, \in \Theta$ let $\leq_{\text{cx}}$ denote the convex order, meaning that $P_{\vtheta_l} \leq_{\text{cx}} P_{\vtheta_u} \iff \bE_{X\sim P_{\vtheta_l}}[\phi(X)] \leq \bE_{Y \sim P_{\vtheta_u}}[\phi(Y)]$ for all convex functions $\phi: \mathcal{Y} \to \Ree$.
In particular for $P_{\vtheta_l} \leq_{\text{cx}} P_{\vtheta_u}$ it holds that $\bE_{X\sim P_{\vtheta_l}}[X] = \bE_{Y\sim P_{\vtheta_u}}[Y]$ and $\bV_{X\sim P_{\vtheta_l}}(X) \leq \bV_{Y\sim P_{\vtheta_u}}(Y)$, since the convex order is a measure of variability of a distribution \citep{book}. 
Similarly, for $Q^\varphi_l, Q^\varphi_u \in \cP(\cP(\cY))$, let $\leq_{\text{cx}}^2$ denote the convex order with respect to all convex functionals $\Phi: \cP(\cY) \to \Ree$.
We define the following properties that a suitable uncertainty measure should satisfy:

\begin{tcolorbox}[colframe=black, colback=gray!2!white, boxsep=0pt, left=3pt, boxrule = 0.9pt]
\begin{itemize}
    \item[A0:] $\tu, \au$, and $\eu$ should be non-negative.
    \item[A1:] $\eu(Q) = 0$ if and only if $Q = \delta_{\vtheta}$.
    \item[A2:] $\eu(\delta_{\vtheta}) \leq \eu(Q_l) \leq \eu(Q_u), \ \forall Q^\varphi_l \leq_{\text{cx}}^2 Q^\varphi_u$.
    \item[A3:] $\au(\delta_{\vtheta_l})\leq \au(\delta_{\vtheta_u}), \ \forall P_{\vtheta_l} \leq_\mathrm{cx} P_{\vtheta_u}$.
    \item[A4:] TU, AU, and EU should be translation invariant.
    \item[A5:] TU, AU, and EU should be scale-order equivariant.
\end{itemize}
\end{tcolorbox}

\emph{Axiom A0.} Non-negativity is a justifiable assumption in the setting of the uncertainty measures, as it allows for interpretability and, together with an additive decomposition, implies that $\tu \geq \eu$ and $\tu \geq \au$.

\emph{Axioms A1 \& A2.} Intuitively, the smallest value of EU should be attained for a distribution with no variability at all; a (second-order) Dirac distribution $\delta_{\vtheta}$. In addition, the converse should hold as well (A1).
Further, when $Q_l^\varphi$ is less variable than $Q_u^\varphi$, as implied by the convex order, the corresponding measure of epistemic uncertainty should assign a smaller value to $Q_l$ (A2). Since the uncertainty measures might depend on $\vtheta$ in a nonlinear way, we assume the convex order on the push-forward distribution $Q^\varphi$.

\emph{Axiom A3.} Similarly, if the second-order distribution is degenerate, and $P_{\vtheta_l}$ is less variable than $P_{\vtheta_u}$, the corresponding measure of $\au$ should assign smaller values as well.

\emph{Axiom A4.} A measure of uncertainty should be invariant to a translation shift in the predictive distribution, i.e., it should not matter whether the first-order distribution is shifted by some constant value. This is formalized in Definition~\ref{def:translation} and should hold for all types of uncertainty.

\emph{Axiom A5.} Finally, all measures of uncertainty should be scale-order equivariant, since a nonnegative scaling of the outcome $Y$ should not affect the order of the assigned uncertainties. Definition~\ref{def:scale_order} formalizes this notion of scale-order equivariant with respect to two scaling functions $g$ and $c$. Note that for the case $g(a) = 1, c(a) = 0$ we obtain \emph{scale equivariance}, which would imply that the assigned uncertainty values do not change at all with a scaling of the outcome variable. While the latter is helpful for interpretability, it might be too strict as a requirement, as it would force uncertainty measures to be unitless\textemdash{}an unrealistic constraint given that uncertainty could naturally change with the underlying unit.

Overall, our axioms are not intended as an exhaustive set of requirements for uncertainty measures in regression, but provide a principled baseline of intuitive properties against which any candidate measure should be evaluated. Yet even this basic set already exposes fundamental differences between the most widely used uncertainty measures.

\section{Comparison of Uncertainty Quantification Methods}
\label{subsec:predictive_models}
In this section, we demonstrate how our uncertainty representation framework includes commonly-used predictive methods, in particular deep ensembles \citep{NIPS2017_9ef2ed4b} and deep evidential regression \citep{amini2020deep}. Furthermore, we highlight how the variance- and entropy-based measures differ across the two representation methods regarding their relative assessment of aleatoric and epistemic uncertainty. More details and derivations of the measures, an analysis of additional representation methods, as well as visualizations are available in \autoref{app:uncertainty_representation}.

\begin{figure}[ht]
\begin{minipage}[c]{0.5\textwidth}
\hspace{3.5em}
Variance-based
\hspace{3.5em}
Entropy-based    
\end{minipage}
\begin{minipage}[c]{0.01\textwidth}
    \phantom{A} \\
    \vspace{-3em}
    \rotatebox{90}{Aleatoric} \\
    \vspace{1.9em}
    \rotatebox{90}{Epistemic}
\end{minipage}
\hspace{0.1em}
\begin{minipage}[c]{0.46\textwidth}
    \centering
    \includegraphics[width = \textwidth]{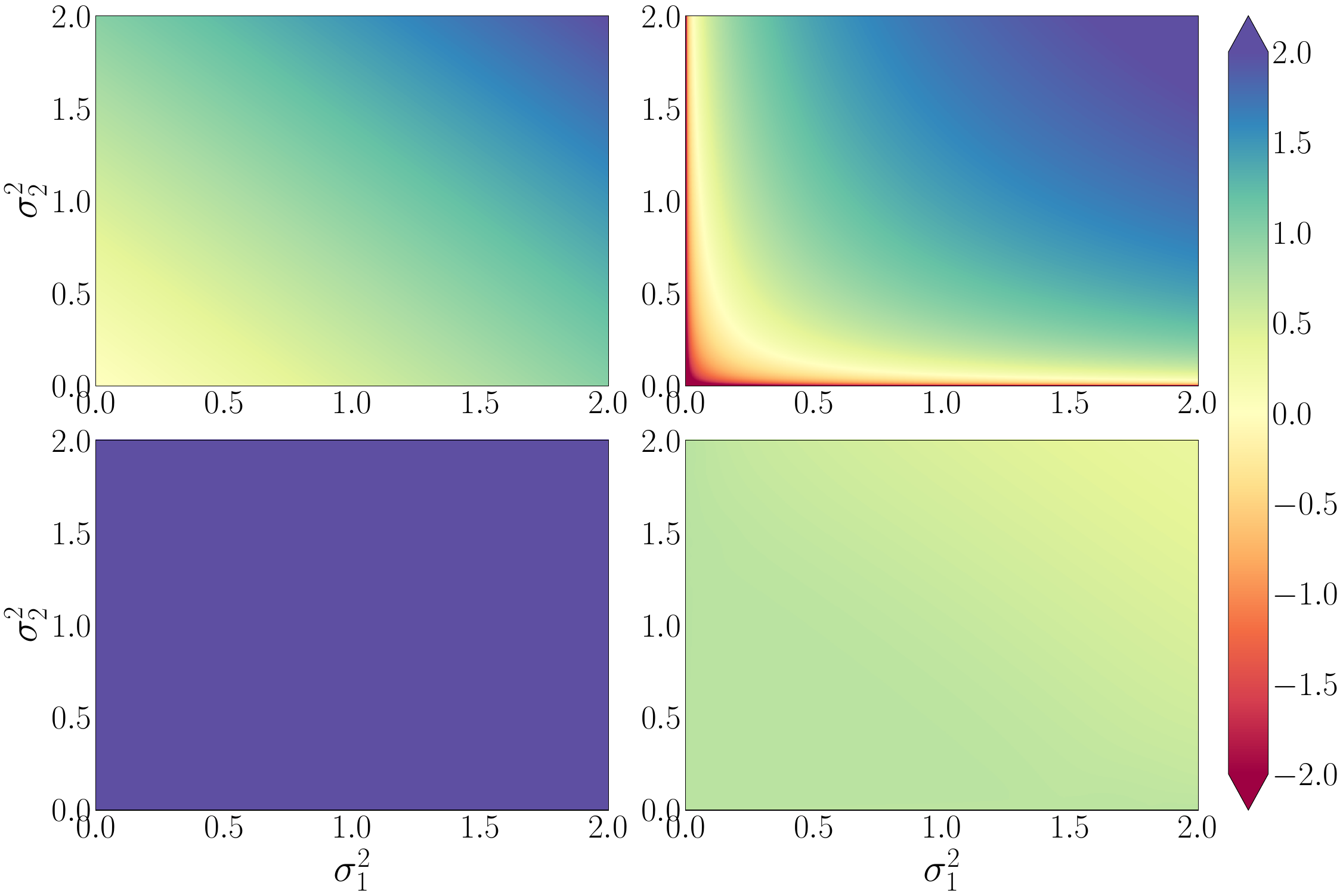} \\
\end{minipage}
\caption{Comparison of the different measures for aleatoric and epistemic uncertainty for a two-member deep ensemble with $\mu_1=2,  \mu_2 = -2$. The variance-based measure for epistemic uncertainty is invariant against changes in $\sigma^2$.}
    \label{fig:ens_au_comparison}
\end{figure}

\emph{Deep Ensembles.} Consider a (univariate) deep ensemble \citep{NIPS2017_9ef2ed4b}, which utilizes a predictive Gaussian distribution, i.e. $P_{\vtheta} = \mathcal{N}(\cdot \,;\mu, \sigma^2), \ \vtheta = (\mu, \sigma^2)^\top \in \Ree \times \Ree_{>0}$. Here, $M$ networks are trained in a stochastic manner, leading to a discrete set of predictions $\vtheta_m = (\mu_m, \sigma_m^2)^\top, \ m=1, \ldots, M$. The ensemble acts as an empirical distribution $Q = \frac{1}{M} \sum_{m=1}^M \delta_{\vtheta_m}$, e.g., a mixture of (equally weighted) Dirac measures over the parameter space. The second-order distribution is then given as $Q^\varphi = \frac{1}{M} \sum_{m=1}^M \delta_{P_{\vtheta_{m}}}$ and the predictive mixture is given as a Gaussian mixture distribution ${\overline{P} =  \frac{1}{M} \sum_{m=1}^M \mathcal{N}(\cdot\, , \mu_m, \sigma_m^2)}$.

For the variance-based measures, we obtain
\begin{align*}
\au(Q) &= \frac{1}{M} \sum_{m=1}^M \sigma_m^2, \\
\eu(Q) &= \frac{1}{M} \sum_{m=1}^M \mu_m^2 - \left( \frac{1}{M}\sum_{m=1}^M \mu_m \right)^2,
\end{align*}
while the entropy-based measures are given as
\begin{align*}
\au(Q) &= \frac{1}{2M} \sum_{m=1}^M \log(2\pi e \sigma_m^2), \\ 
\eu(Q) &= \frac{1}{M} \sum_{m=1}^M \int_{\Ree} \phi(t \mid \vtheta_m) \log \frac{\phi(t
 \mid \vtheta_m) }{\frac{1}{M}\sum_{l=1}^M \phi(t \mid \vtheta_l) }\, dt,
\end{align*}
where $\phi(\cdot \mid \vtheta_m)$ denotes the density of a Gaussian with parameters $\vtheta_m$.

Clearly, both measures behave quite differently. In particular, the entropy-based measure does not admit a closed-form expression for epistemic uncertainty.
For the variance-based measure, EU depends only on the first moment and AU only on the second moment of the underlying distribution. Therefore, the measure cannot distinguish between distributions where only higher-order moments are different, leading to a violation of (A1), as highlighted in \autoref{fig:ens_au_comparison}.

\emph{Deep Evidential Regression.} Consider the deep evidential regression framework \citep{amini2020deep}, which also assumes a predictive Gaussian distribution. Here, second-order uncertainty is incorporated by specifying a Normal Inverse-Gamma (NIG) prior over the Gaussian parameters $\vtheta = (\mu, \sigma^2)^\top$. Applying this to our setting, we obtain: $Q= q(\mu, \sigma^2) = q(\mu) \, q(\sigma^2) = \mathcal{N}(\mu; \gamma, \sigma^2 \upsilon^{-1}) \, \Gamma^{-1} (\sigma^2; \alpha, \beta)$ with $\gamma \in \Ree, \upsilon>0, \alpha >1, \beta >0$.
While the push-forward $Q^\varphi$ is only defined implicitly with no analytic expression, the predictive mixture is given via a Student's t distribution $\overline{P} = t_{2\alpha}\left(\cdot \, ; \gamma, \frac{\beta(1+\upsilon)}{\alpha \upsilon} \right)$ with $2\alpha$ degrees of freedom.

\begin{figure}[ht]
\centering
\begin{subfigure}{\linewidth}
\centering
\caption{$\alpha$-evidence}
\includegraphics[width = \linewidth]{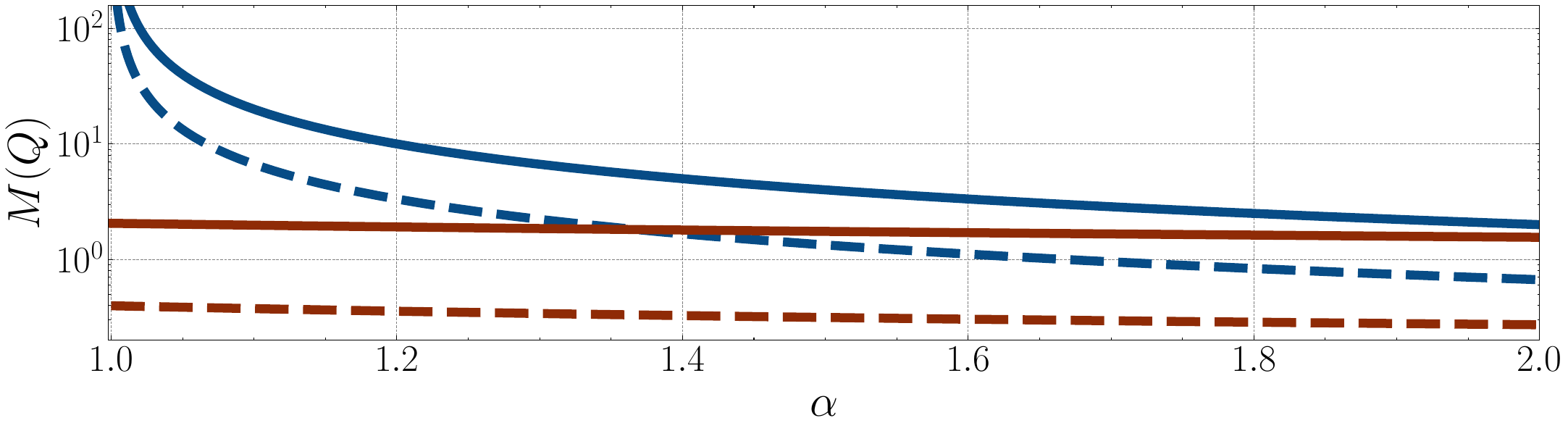}
\end{subfigure}
\begin{subfigure}{\linewidth}
\centering
\caption{$\upsilon$-evidence}
\includegraphics[width = \linewidth]{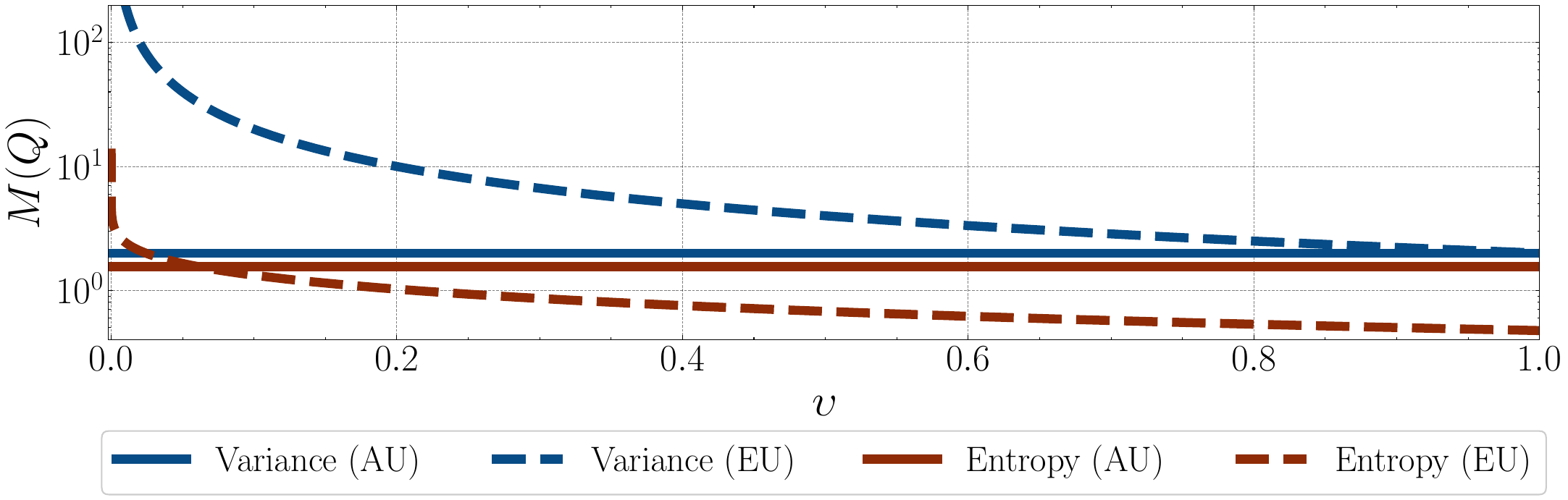}
\end{subfigure}
\caption{Comparison of the asymptotic behavior of the deep evidential regression parameters $\alpha, \upsilon$ under the two different uncertainty measures. Although both measures behave similarly for $\upsilon$, the variance-based measures diverge for $\alpha \downarrow 1$, while the entropy-based measure remains bounded.}
\label{fig:der_asymptotics}
\end{figure}

In the deep evidential regression setting, the variance-based measures are given as
%\vspace{-0.2cm}
\begin{align*}
    \au(Q) &=  \frac{\beta}{\alpha - 1}, \qquad
    \eu(Q) =  \frac{\beta}{\upsilon(\alpha - 1)},
\end{align*}
while for the entropy-based measures, we obtain
\begin{align*}
    \au(Q) &= \frac{1}{2} \left(\log(2\pi e)+\log(\beta)- \psi(\alpha)\right), \\
     \eu(Q) &= H\left(t_{2\alpha}\left(\gamma, \frac{\beta(1+ \upsilon)}{\alpha \upsilon}\right)\right) - \au(Q),
\end{align*}    
where $\psi$ denotes the digamma function. Again, both measures differ significantly in their behavior with respect to the parameters of the second-order distribution. In particular, as visualized in \autoref{fig:der_asymptotics}, both measures behave very differently for decreasing parameters $\alpha, \upsilon$. Intuitively, for the NIG distribution, the mean can be interpreted as being estimated from $\upsilon$ virtual observations, while its variance is estimated from $\alpha$ virtual observations \citep{amini2020deep}. Therefore, the limiting cases $\alpha \downarrow 1$ and $\upsilon \downarrow 0$ describe the behavior of the model under decreasing evidence. While the entropy-based measures remain well-defined for $\alpha \downarrow 1$, the variance-based measure diverges. For $\upsilon$, both measures behave similarly, with AU remaining bounded and EU diverging, since with zero evidence $\upsilon$, there is maximal uncertainty about the model parameters. This directly relates to the existence assumption of the measures; since the variance based measures requires finite second-moments, it diverges in the limiting case. 
Further visualizations of the behavior of the different measures for varying $\alpha, \beta$ are available in \autoref{app:uncertainty_representation}. 

\section{Assessment of Properties}
In this section, we demonstrate how our proposed axioms can be used to assess existing or novel uncertainty measures theoretically and can ultimately guide practitioners in selecting or designing measures with specific characteristics, depending on the underlying task.
Here, we specifically analyze the previously introduced variance- and entropy-based measures, with respect to the proposed axioms. As already highlighted in \autoref{fig:ensemble_comparison} and \autoref{fig:ens_au_comparison}, the same prediction can lead to substantially different estimations of uncertainties. This also transfers to the axioms, where we prove that both measures fulfill, but also violate some of the axioms, highlighting individual strengths and weaknesses. All corresponding proofs can be found in \autoref{app:proofs}.

\begin{proposition}
\label{prop:var_a1}
    The variance-based measures violate A1.
\end{proposition}
The variance-based measure can only measure epistemic uncertainty in terms of the mean of the predictive distribution, and therefore cannot distinguish between two different second-order distributions that have the same marginal distribution over the mean. For instance, for a predictive Gaussian $p(y \mid \vvtheta \,) = \mathcal{N}(\mu, \sigma^2)$, the epistemic uncertainty reduces to $\eu(Q) = \bV_Q(\mu)$ and is therefore insensitive to the marginal distribution of $\sigma^2$. Consequently, two second-order distributions that share the same marginal over $\mu$ but differ in their distribution over $\sigma^2$ are indistinguishable—compare the deep-ensemble example in \autoref{fig:ens_au_comparison}.

\begin{proposition}
\label{prop:var_fulfills}
    The variance-based measures fulfill A0, A2, A3, A4, A5.
\end{proposition}
By nonnegativity of the variance, A0 follows immediately. A2 and A3 follow from the connection between convex order and increasing variance. Furthermore, A4 also holds, as the variance is translation-invariant. Finally, the variance-based measures are scale-order equivariant with $g(a) = a^2$ and $c(a) = 0$. Therefore, the variance-based measures fulfill the majority of the proposed axioms.

\begin{proposition}
\label{prop:entropy_a0}
    The entropy-based measure violates A0 for AU and TU.
\end{proposition}
As highlighted in \autoref{fig:ensemble_comparison}, one major drawback of the differential entropy and the entropy-based measure is that although EU is nonnegative, AU (and therefore TU) can be assigned negative values. A natural definition of the absence of uncertainty is to have zero uncertainty, or in other words, complete knowledge about the underlying process. Negative uncertainty, on the other hand, does not admit a natural interpretation and makes the entropy-based measure difficult to evaluate in practice and impossible to compare to other measures. Since the lower bound for AU regarding entropy is negative infinity, this problem cannot be adjusted through a linear transformation. Furthermore, this implies that TU is not necessarily larger than EU, which is counterintuitive regarding the very definitions of the underlying uncertainties.

\begin{figure}[t]
\centering
\includegraphics[width = \linewidth]{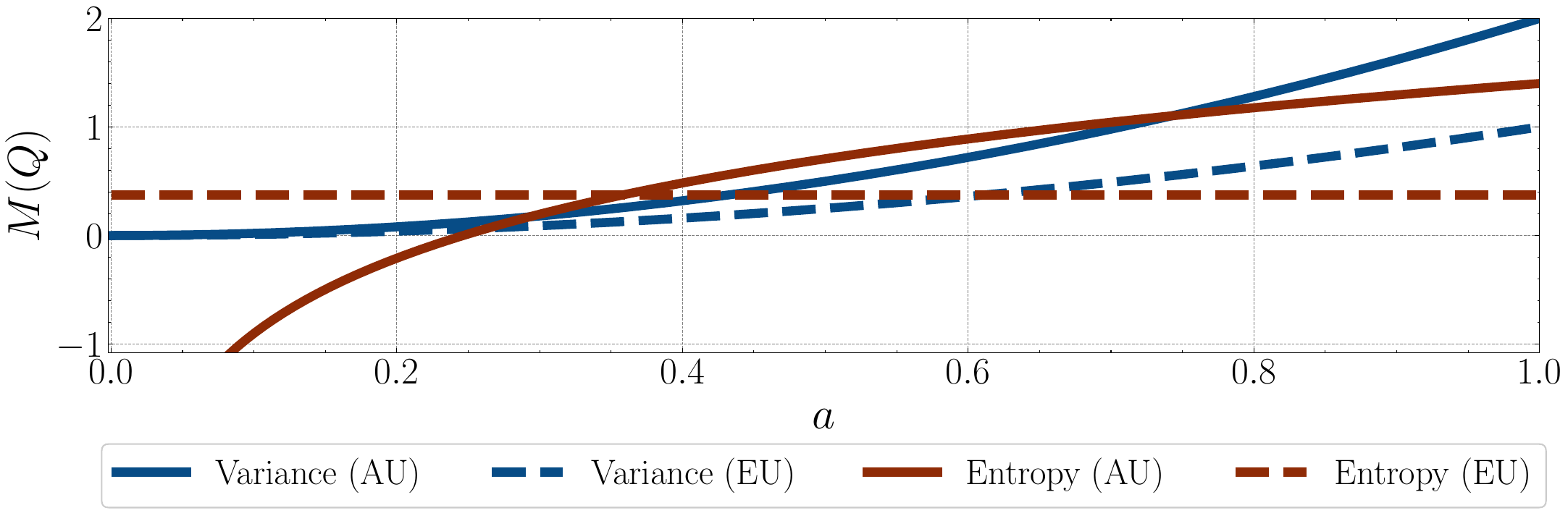}
\caption{Comparison of the scale-order equivariance for the case of univariate deep evidential regression. While each of the entropy-based measures is equivariant when scaling the underlying data with a factor $a$, their ordering depends on the chosen scaling. In particular, AU is smaller (larger) than EU for $a \approx 0$ ($a\approx 1$), which can lead to problems as the ordering of the uncertainties should not depend on the scaling of the data. For the variance-based measure this problem does not occur.}
\label{fig:scale_order_asymptotics}
\end{figure}

\begin{proposition}
\label{prop:entropy_fulfill}
    The entropy-based measure fulfills A2, A4, and A5.
\end{proposition}
A2 follows from the definition of convex order and convexity of the Kullback-Leibler divergence in $P$. A4 follows since all involved quantities are translation-invariant. Finally, A5 follows by properties of the differential entropy with $g(a) = 1,\ c(a) = d\log(a)$ for TU, AU and by properties of the KL-divergence with $g(a) = 1, \ c(a) = 0$ for EU.
Interestingly, while the entropy-based measures fulfill scale-order equivariance for each individual measure, as the shift function $c$ is different across EU and AU, their ordering changes with the scaling factor $a$. In fact, since EU is scale equivariant and nonnegative, it is always possible to find $a_l, a_u > 0$ such that $\au(Q^{a_l}) < \eu(Q^{a_l}) = \eu(Q^{a_u}) \leq \au(Q^{a_u})$, as visualized in \autoref{fig:scale_order_asymptotics}. From an interpretability perspective, this is problematic, as the scaling of the data impacts the connection of EU and AU, in particular, which of the two is larger. This is directly related to the entropy-based measure violating (A0).

\begin{proposition}
\label{prop:entropy_partial}
    The entropy-based measure fulfills A1 if the parametric family is identifiable and A3 if the parametric family is log-concave.
\end{proposition}
In general, the differential entropy and KL-divergence can depend on the parameter $\vtheta$ in any complex and nonlinear way. Therefore, for these axioms to hold, additional assumptions are required. A1 holds if the parametric family is identifiable, which is a reasonable and commonly used assumption. A3 holds if the parametric family is log-concave, which includes distributions such as the normal or Gamma distribution. Both assumptions are fulfilled for the previously introduced uncertainty representation methods. However, for both axioms, it can be shown that they do not hold without the additional assumptions discussed above.

\section{Related work}
Many works focus on analyzing different measures of uncertainty and their corresponding properties in the classification setting. Specifically, axiomatic evaluation has been put forward in the machine learning community \citep{hullermeier2022quantification} and built the foundation for the development of many new measures of uncertainty, for example, based on proper scoring rules \citep{kotelevskii2025from,hofman2024quantifying}, Wasserstein distance \citep{sale2023second} or credal sets \citep{hofman2024quantifying}. Despite this focus on an axiomatic understanding of uncertainty measures for classification, in the regression case, research mainly focuses on uncertainty representation and only little on a formal understanding of uncertainty measures and their underlying properties.

\cite{amini2020deep, meinert2022multivariatedeepevidentialregression} propose the (multivariate) deep evidential regression, as a generalization of the evidential learning framework. To assess aleatoric and epistemic uncertainty, they utilize the variance-based decomposition. \cite{laves2021recalibrationaleatoricepistemicregression} propose a calibration method that removes bias from an estimate of aleatoric uncertainty by correcting the scale of the first-order predictive variance. They also utilize the variance-based measure to decompose the different types of uncertainty. \cite{valdenegrotoro2022deeperlookaleatoricepistemic} compare several different methods for uncertainty representation, such as deep ensembles or dropout, with respect to their capacity to measure and decompose epistemic and aleatoric uncertainty. For their assessment, they utilize a first-order predictive Gaussian and the variance-based uncertainty decomposition. For the selected methods, they critique the lack of disentanglement between aleatoric and epistemic uncertainty across different prediction tasks.

Moving to entropy-based methods, \cite{berry2024efficientepistemicuncertaintyestimation} propose a combination of deep ensembles and normalizing flows to generate a non-parametric model to estimate uncertainties. In their approach, aleatoric uncertainty is measured via conditional entropy with respect to the output distribution of the normalizing flow. Furthermore, they provide so-called Pairwise-Distance Estimators to provide bounds on the differential entropy and therefore allow for representing the epistemic uncertainty in the base distribution of the normalizing flow in an analytical way. \cite{malinin2020regressionpriornetworks} introduce regression prior networks, which are similar to multivariate deep evidential regression \citep{meinert2022multivariatedeepevidentialregression} and utilize a prior distribution over the parameters of a predictive multivariate Gaussian. To assess uncertainties, they utilize the variance- and entropy-based decomposition, as well as an additional measure called the expected pairwise KL-divergence, which is an upper bound on mutual information and is analytically available for a larger class of models.
% The regression prior networks exhibit superior performance on single-model, as well as ensemble approaches. 
\cite{postels2021hiddenuncertaintyneuralnetworks} consider an approach that utilizes a latent representation of a neural network to measure epistemic and aleatoric uncertainty. By decomposing the network into a latent representation for each of the $L$ layers, they can utilize bounds on differential entropy to provide estimates for both types of uncertainty. Using a quite different approach, \cite{lahlou2023deupdirectepistemicuncertainty} propose to characterize epistemic and aleatoric uncertainty in terms of the Bayes risk on an estimator. By utilizing out-of-sample errors of a trained model, they obtain an estimate for the excess risk, which represents epistemic uncertainty. For aleatoric uncertainty, they propose several heuristics, such as estimation based on intrinsic data variation.

While all of the aforementioned approaches provide novel and well-founded ways to measure epistemic and aleatoric uncertainty, the focus is usually set on a specific predictive model and corresponding uncertainty representation. To the best of our knowledge, there exists no work that considers a general uncertainty representation framework, formalizes the corresponding evaluation procedure in a theoretical and axiomatic manner in general regression tasks, and compares the properties of different measures in a theoretical and axiomatic manner.

\section{Discussion} 
This work establishes a formal foundation for uncertainty representation, quantification, and evaluation in the supervised regression setting. We introduce a parametric uncertainty representation framework that allows for operationalization via a tractable second-order distribution, induced by the distribution over the parameter space and encompassing many well-known predictive machine learning methods, thereby enabling the first systematic cross-method comparison in regression.
Furthermore, we develop an axiomatic methodology, comparing both classification-adapted and regression-specific axioms, that allows for analyzing given uncertainty measures with respect to their fundamental properties, strengths, and weaknesses.
We demonstrate the practicality of our approach by evaluating the commonly used entropy- and variance-based uncertainty measures, exposing the strengths and weaknesses of each. In particular, we show that entropy-based measures can yield negative values in certain settings, which directly impacts scale equivariance across AU and EU and hinders natural interpretability.
In contrast, the variance-based approach satisfies the majority of the proposed axioms but lacks a unique minimum for epistemic uncertainty, rendering it insensitive to uncertainty in higher-order moments, and requires stricter assumptions for its existence.
Beyond the specific measures analyzed here, our axiomatic framework provides practitioners and researchers with a principled methodology for evaluating existing uncertainty measures and provides a principled guideline for designing novel uncertainty measures.

\paragraph{Limitations}
While choosing a predictive model with uncertainty only in its parameters is very general and includes commonly used approaches, as highlighted in this paper, it might be too restrictive in certain settings, for example, when Gaussian processes or sample-based generative models, such as diffusion models.
In addition, although the proposed axioms provide a principled baseline, the selection is not exhaustive; alternative or additional axioms, specifically designed for the regression domain, could reveal further distinctions between uncertainty measures not captured here.
Finally, while the axioms capture theoretically desirable properties, satisfying or violating them does not directly guarantee good or poor empirical performance in any particular downstream task. The relationship between axiomatic compliance and practical utility\textemdash{}for example, in uncertainty-guided decision making or out-of-distribution detection\textemdash{}remains an open question.

\paragraph{Future work}
Our work opens several venues for future research. 
The most immediate open problem is to search for novel uncertainty measures that satisfy the proposed axioms in their entirety\textemdash{}or to establish whether such measures exist in the first place. 
For the classification setting, new measures have been proposed based on Wasserstein distance \citep{sale2023second} or score-based entropies \cite{kotelevskii2025from,hofman2024quantifying}, and an extension to the regression setting, as well as a corresponding theoretical analysis, seems promising. Here, our proposed axioms can provide a clear guidance on the design of new measures, for example, measures based on statistical divergences seem to be a natural candidate to fulfill (A1) and (A2).
In addition, one may think of further analyzing and extending the set of axioms itself to accommodate other desirable properties of uncertainty measures, such as robustness or interpretability.
Furthermore, validating the axioms in real-world regression settings could lead to a deeper understanding of the strengths and weaknesses of individual approaches. In particular, it would be worth analyzing whether the implications of the axioms persist across different data modalities.
Finally, generalizing the framework to a more general class of predictive distributions is a direction worth pursuing. This could allow for analyzing the axioms for arbitrary predictive distributions, such as nonparametric methods or generative models.

\begin{acknowledgements}
C. Bülte, Y. Sale, G. Kutyniok, and E. H\"ullermeier acknowledge support by the DAAD programme Konrad Zuse Schools of Excellence in Artificial Intelligence, sponsored by the Federal Ministry of Research, Technology and Space.

C. Bülte and G. Kutyniok acknowledge support by the German Research Foundation under the grant DFG-SPP-2298.

E. H\"ullermeier acknowledges support by the German Research Foundation under the grant GRK 3081 (project number 534429653).

G. Kutyniok also acknowledges support by the gAIn project, which is funded by the Bavarian Ministry of Science and the Arts (StMWK Bayern) and the Saxon Ministry for Science, Culture and Tourism (SMWK Sachsen). Furthermore, G. Kutyniok is supported by LMUexcellent, funded by the Federal Ministry of Education and Research (BMBF) and the Free State of Bavaria under the Excellence Strategy of the Federal Government and the Länder as well as by the Hightech Agenda Bavaria.
\end{acknowledgements}

\bibliography{references}

\newpage

\onecolumn

\title{An Axiomatic Assessment of Entropy- and Variance-based Uncertainty Quantification in Regression\\(Supplementary Material)}
\maketitle
\vspace{0.5cm}

\appendix

\section{Proofs of Propositions}
\label{app:proofs}
\subsection{Variance-based}

\begin{proof}[Proof of Proposition~\ref{prop:var_a1}]
We prove that the variance-based measure violates A1.
    Consider $P_1 = \mathcal{N}(0, \sigma_1^2), P_2 = \mathcal{N}(0, \sigma_2^2)$ with $\sigma_1^2 \neq \sigma_2^2$ and a second-order distribution, specified as a Dirac mixture, i.e. $Q = \frac{1}{2} \delta_{\sigma_1^2} + \frac{1}{2} \delta_{\sigma_2^2}$. Recall that for the variance-based measure, we have $\eu(Q) = \bV_Q[\bE[Y \mid \vvtheta]]$, with $\vvtheta \sim Q$. In addition, we obtain $\overline{P} = \frac{1}{2} P_1 + \frac{1}{2} P_2$ and $\bE_{Y' \sim \overline{P}}[Y'] = 0$. Then we obtain
\begin{align*}
    \eu(Q) & = \bV_Q[\bE[Y \mid \vvtheta]]=  \bE_{\vvtheta \sim Q}[(\bE_{Y \sim P_{\vvtheta}}[Y] - \underbrace{\bE_{Y' \sim \overline{P}}[Y']}_{=0})^2 ] =\bE_{\vvtheta \sim Q}[( \bE_{Y \sim P_{\vvtheta}}[Y])^2 ] \\
           & = \frac{1}{2}(\underbrace{\bE_{Y \sim P_1}[Y]}_{=0})^2  + \frac{1}{2}( \underbrace{\bE_{Y \sim P_2}[Y]}_{=0})^2 = 0.
\end{align*}
Therefore, we obtain $\eu(Q) = 0$ although $Q \neq \delta_{\vtheta}$. 
\end{proof}

\begin{proof}[Proof of Proposition~\ref{prop:var_fulfills}]
We prove that the variance-based measures fulfill A0, A2, A3, A4, and A5.

\emph{A0: }
A0 follows immediately from the variance being nonnegative.

\emph{A2: }
By definition of the convex order and the push-forward measure, we have
\[
\bE_{Q_l^\varphi} [\Phi(P)] = \bE_{\vvtheta \sim Q_l}[\Phi(P_{\vvtheta})] \leq_\mathrm{cx}^2 \bE_{\vvtheta \sim Q_u}[\Phi(P_{\vvtheta})] = \bE_{Q_u^\varphi} [\Phi(P)],
\]
for all convex functionals $\Phi: \cP(\cY) \to \Ree$. By definition, convex order preserves expectations, which implies equal predictive mixtures, i.e., $\overline{P}_{Q_l} = \overline{P}_{Q_u} \coloneq \overline{P}$.

Define the random vectors $\boldsymbol{M}_l \coloneq \bE[\bY \mid \vvtheta], \ \vvtheta \sim Q_l$ and $\boldsymbol{M}_u \coloneq \bE[\bY \mid \vvtheta], \ \vvtheta \sim Q_u$. Then, for the variance-based measure, we have $\eu(Q_l) = \bV(\boldsymbol{M}_l), \quad \eu(Q_u) = \bV(\boldsymbol{M}_u).$

Now, consider the functional $\Phi(P) \coloneq g(\bE_{\bY \sim P}[\bY])$, which is convex on $\cP(\cY)$ for any convex $g: \Ree^{d} \to \Ree$. Using the definition of the convex order, we obtain
\[
\bE_{\vvtheta \sim Q_l}[g(\bE_{\bY \sim P_{\vvtheta}})] \leq \bE_{\vvtheta \sim Q_u}[g(\bE_{\bY \sim P_{\vvtheta}})] \quad \forall \ \mathrm{convex} \ g,
\]
which, by definition, is
\[
\bE_{Q_l}[g(\boldsymbol{M}_l)] \leq \bE_{Q_u}[g(\boldsymbol{M}_u)] \quad \forall \ \mathrm{convex} \ g,
\]
and implies $\boldsymbol{M}_l \leq_\mathrm{cx} \boldsymbol{M}_u$\footnote{Here we slightly overload the notation and use $\leq_\mathrm{cx}$ for the convex order, as introduced earlier, but comparing random vectors instead of probability measures.}. From the convex order it follows immediately that $\bE[\boldsymbol{M}_l] = \bE[\boldsymbol{M}_u]$ and $\bV(\boldsymbol{M}_l) \leq \bV(\boldsymbol{M}_u)$, which directly implies
\[
\eu(Q_l) \leq  \eu(Q_u).
\]

\emph{A3: }
Recall that we have $P_{\vtheta_l} \leq_\mathrm{cx} P_{\vtheta_u}, \ \vtheta_l,\vtheta_u \in \Theta$ and $\au(\delta_{\vtheta}) = \bE_{\delta_{\vtheta}}[\bV(P_{\vtheta})] = \bV(P_{\vtheta}).$ By definition of the convex order we have
\[
P_{\vtheta_l} \leq_\mathrm{cx} P_{\vtheta_u} \implies \bV(P_{\vtheta_l}) \leq \bV(P_{\vtheta_u}),
\]
and therefore
\[
\au(\delta_{\vtheta_l}) \leq \au(\delta_{\vtheta_u}).
\]

\emph{A4: }
Let $Y \mid \vvtheta \sim P_{\vvtheta}$ and $Y_c \mid \vvtheta' \sim P_{\vvtheta'}$ with $\vvtheta' = \tau_c(\vvtheta)$ with $\vvtheta \sim Q$. Then, by definition, we have $Y \overset{d}{=}Y+c$. By the properties of the variance, we obtain
\[
\au(Q_c) = \bE_{Q_c}[\bV(Y \mid \vvtheta')] = \bE_{Q}[\bV(Y+c \mid \tau_c(\vvtheta))] = \bE_{Q}[\bV(Y \mid \vvtheta)] = \au(Q),
\]
and
\[
\eu(Q_c) = \bV_{Q_c}(\bE[Y \mid \vvtheta']) =  \bV_{Q}(\bE[Y+c \mid \tau_c(\vvtheta)]) =  \bV_{Q}(\bE[Y \mid \vvtheta]) = \eu(Q).
\]
TU then follows from the additive decomposition.

\emph{A5: }
By definition, we have $P_{\tau_a(\vvtheta)} = S_a P_{\vvtheta}$ for $\vvtheta \sim Q$. If $Y \mid \vvtheta \sim P_{\vvtheta}$, then $Y^a \mid \vvtheta' \sim S_a P_{\vvtheta'}$ is the law of $aY$ with $\vvtheta' = \tau_a(\vvtheta) \sim Q^a$. By properties of the variance, we obtain
\[
\au(Q^a) = \bE_{Q^a}[\bV(Y \mid \vvtheta')] = \bE_{Q}[\bV(Y^a \mid \vvtheta')] = \bE_{Q}[a^2\bV(Y \mid \vvtheta)] = a^2 \au(Q),
\]
and
\[
\eu(Q^a) = \bV_{Q^a}(\bE[Y \mid \vvtheta']) = \bV_{Q}(\bE[Y^a \mid \vvtheta']) = \bV_{Q}(a \bE[Y \mid \vvtheta]) = a^2\eu(Q).
\]
TU follows from the additive decomposition and we obtain $g(a) = a^2$ and $c(a) = 0$.
\end{proof}

\subsection{Entropy-based}
\begin{proof}[Proof of Proposition~\ref{prop:entropy_a0}]
We prove that the entropy-based measure violates A0 for AU and TU.
For EU, as the KL-divergence is nonnegative by definition, EU is also nonnegative. AU can be negative since differential entropy can be negative. As an example, consider the exponential distribution with entropy $H(\text{EXP}(\lambda)) = 1 - \log(\lambda)$, which is negative for $\lambda > e$. Taking $Q = \delta_{2e}$ as a second-order distribution then leads to $\au(Q) <0$. Due to its additive structure, total uncertainty $\tu = \eu + \au$ can also be negative.
\end{proof}

\begin{proof}[Proof of Proposition~\ref{prop:entropy_fulfill}]
We prove that the entropy-based measures fulfill A2, A4, and A5.

\emph{A2: }
By definition of the convex order and the push-forward measure, we have
\[
\bE_{Q_l^\varphi} [\Phi(P)] = \bE_{\vvtheta \sim Q_l}[\Phi(P_{\vvtheta})] \leq_\mathrm{cx}^2 \bE_{\vvtheta \sim Q_u}[\Phi(P_{\vvtheta})] = \bE_{Q_u^\varphi} [\Phi(P)],
\]
for all convex functionals $\Phi: \cP(\cY) \to \Ree$. By definition, convex order preserves expectations, which implies equal predictive mixtures, i.e., $\overline{P}_{Q_l} = \overline{P}_{Q_u} \coloneq \overline{P}$.

For the entropy-based measure we have $\eu(Q) = \bE_Q[\Phi(P_{\vvtheta})]$, where $\Phi(P) \coloneq D_\mathrm{KL}(P \| \overline{P})$ is a convex functional, due to the convexity of the KL-divergence in the first argument. Therefore, by the assumption of $Q_l^\varphi \leq_\mathrm{cx}^2 Q_u^\varphi$, we obtain
\[
\eu(Q_l) = \bE_{\vvtheta \sim Q_l}[D_\mathrm{KL}(P_{\vvtheta} \mid \overline{P})] \leq \bE_{\vvtheta \sim Q_u}[D_\mathrm{KL}(P_{\vvtheta} \mid \overline{P})] = \eu(Q_u).
\]

\emph{A4: }
Let $\vvtheta' = \tau_c(\vvtheta)$ with $\vvtheta \sim Q$.
By the shift invariance of the differential entropy $H(\cdot)$, we obtain
\[
\au(Q_c) = \bE_{Q_c}[H(P_{\vvtheta'})] = \bE_Q[H(P_{\tau_c(\vvtheta)})] = \bE_Q[H(P_{\vvtheta}(\cdot - c))] = \bE_Q[H(P_{\vvtheta})] = \au(Q),
\]
and
\begin{align*}
\eu(Q_c) &= H(\bE_{Q_c}[P_{\vvtheta'}]) - \au(Q_c) = H(\bE_Q[P_{\vvtheta}(\cdot - c)])  - \au(Q)\\ &= H(\overline{P}_{\vvtheta}(\cdot - c)) -\au(Q) = H(\overline{P}_{\vvtheta}) -\au(Q) = \eu(Q).
\end{align*}
TU then follows from the additive decomposition.

\emph{A5: }
By definition, we have $P_{\tau_a(\vvtheta)} = S_a P_{\vvtheta}$ for $\vvtheta \sim Q$. If $Y \mid \vvtheta \sim P_{\vvtheta}$, then $Y^a \mid \vvtheta' \sim S_a P_{\vvtheta'}$ is the law of $aY$ with $\vvtheta' = \tau_a(\vvtheta) \sim Q^a$.
By properties of the differential entropy $H(\cdot)$, we obtain
\[
\au(Q^a) = \bE_{Q^a}[H(P_{\vvtheta'})] = \bE_{Q^a}[H(S_a P_{\vvtheta})] = \bE_{Q^a}[H(P_{\vvtheta}) + d\log a] = \au(Q) + d\log a.
\]
Furthermore, we have
\[H(\bE_{Q^a}[P_{\vvtheta'}]) = H(\bE_{Q}[P_{\tau_a(\vvtheta)}]) = H(\bE_{Q}[S_a P_{\vvtheta}]) = H(S_a \overline{P}) = H(\overline{P}) + d\log a,
\]
and therefore
\[
\eu(Q^a) = H(\bE_{Q^a}[P_{\vvtheta'}]) - \au(Q^a) = (H(\overline{P}) + d\log a) - (\au(Q) + d\log a) = \eu(Q).
\]
With this, we obtain $g(a) = 1$, $c(a) = d\log a$ for AU, TU and $c(a) = 0$ for EU.
\end{proof}

\begin{proof}[Proof of Proposition~\ref{prop:entropy_partial}]
Here, we prove that the entropy-based measure fulfills A1 if the parametric family is identifiable and A3 if the parametric family is log-concave.

First, we show that the entropy-based measure fulfills A1 if the parametric family is identifiable.

Let $Q \in \mathcal{P}(\Theta), \bar{P} = \int P_{\vtheta}\,dQ(\vtheta)$, and define $Q^\varphi = \phi_{\#}Q \in \mathcal{P}(\mathcal{P}(\mathcal{Y}))$ as the push-forward of $Q$ under $\varphi$. Then
$$\eu(Q)= \bE_Q[D_\mathrm{KL}(P_{\vvtheta}\mid \overline{P})] = \int D_{\mathrm{KL}}(P\|\overline{P})\,Q^\varphi(dP)\geq 0.$$
In particular, we obtain $\eu(Q)=0$ iff $Q^\varphi=\delta_{\bar{P}}$.
Under identifiability of $\varphi$, this is equivalent to $Q = \delta_{\vtheta}$.

Second, we show that the entropy-based measure fulfills A3 if the parametric family is log-concave.
  A probability distribution has log-concave density if the density can be expressed as $p(x) \equiv \exp(\varphi(x))$ for a concave function $\varphi(x)$. Since by assumption we have $P_{\vtheta_l} \leq_{\text{cx}} P_{\vtheta_u}$ and $p_{\vtheta_u}$ is log-concave, one obtains $\mathrm{supp}(P_{\vtheta_l}) \subseteq \mathrm{supp}(P_{\vtheta_u})$ \citep{10.1214/aoms/1177700153} such that the cross-entropy defined below is finite.
  Now, recall that differential entropy, which can be expressed as
    \begin{align*}
        \au(\delta_{\vtheta}) = H(P_{\vtheta})\coloneq - \int p_{\vtheta}(x) \log  p_{\vtheta}(x) d\mu(x) = \bE_{P_{\vtheta}}[-\log p_{\vtheta}(X)].
    \end{align*}
    Then, for a log-concave density, we have that $\phi(x) \coloneq -\log p_{\vtheta_u}(x)$ is a convex function in $x$. By convex order, we then have
    \[
    \bE_{X \sim P_{\vtheta_l}}[- \log p_{\vtheta_u}(X)] = \bE_{X \sim P_{\vtheta_l}}[\phi(X)] \leq \bE_{Y \sim P_{\vtheta_u}}[\phi(Y)] = \au(_{\vtheta_u}).
    \]
    The left-hand side is the cross-entropy of $P_{\vtheta_l}, P_{\vtheta_u}$, which, by definition, can be decomposed into 
    \[
  \bE_{X \sim P_{\vtheta_l}}[- \log p_{\vtheta_u}(X)] = H(P_{\vtheta_l}) + D_\mathrm{KL}(P_{\vtheta_l} \| P_{\vtheta_u}) \geq H(P_{\vtheta_l}),
    \]
    where the inequality follows from the KL-divergence being nonnegative. Combining the above gives
    \[
     \au(\delta_{{\vtheta_l}}) \leq \bE_{X \sim P_{\vtheta_l}}[- \log p_{\vtheta_u}(X)] = \bE_{X \sim P_{\vtheta_l}}[\phi(X)] \leq \bE_{Y \sim P_{\vtheta_u}}[\phi(Y)] = \au(\delta_{\vtheta_u}).
    \]    
\end{proof}

\begin{proposition}
   The entropy-based measure violates A1 if the model is not identifiable.
\end{proposition}
\begin{proof}
Consider a non-identifiable model $P_{\vtheta}$ with $P_{\vtheta_1} = P_{\vtheta_2}$ and $\vtheta_1 \neq \vtheta_2$. Then $Q=\frac{1}{2}\delta_{\vtheta_1}+\frac{1}{2}\delta_{\vtheta_2}$ is not the Dirac measure, but we obtain $\overline{P} = \frac{1}{2}P_{\vtheta_1} + \frac{1}{2}P_{\vtheta_2} = P_{\vtheta_1} = P_{\vtheta_2}$, and therefore
\[
\eu(Q) = \bE_{\vvtheta \sim Q}[D_\mathrm{KL}(P_{\vvtheta}\mid \overline{P})] = 0.
\]    
\end{proof}

\begin{proposition}
    Without additional assumptions, the entropy-based measure violates A3.
\end{proposition}
\begin{proof}
Consider the (symmetric) Beta distribution $Y_\alpha \sim P_{\alpha} = \mathrm{Beta}(\alpha, \alpha), \ \alpha > 0$, which has mean $\bE_{P_\alpha}[Y_\alpha] = 1/2$ and a second-order Dirac distribution $Q = \delta_\alpha$. Consider two special cases:
\begin{itemize}
    \item $Y_1 \sim \mathrm{Beta}(1,1) = \mathrm{Unif}(0,1)$ with CDF $P_1(x) = x, \, x \in [0,1]$.
    \item $Y_{1/2} = \mathrm{Beta}(1/2,1/2)$ (arcsine distribution) with cdf $P_{1/2}(x) = \frac{2}{\pi} \mathrm{arcsin}\left( \sqrt{x} \right), \, x \in (0,1)$.
\end{itemize}

First, we check the convex order. From \cite[Theorem 3.A.5]{book} we know that for two variables $X,Y$ with equal means and CDF $F$ and $G$, respectively, we have
\[
X \leq_\mathrm{cx}Y \iff \int_0^p F^{-1}(u) \, du \geq \int_0^p G^{-1}(u) \, du ,\quad \forall p \in [0,1].
\]
In our case, we have
\[
\int_0^p P_{1}^{-1}(u) \, du = \int_0^p u \, du \geq \int_0^p \sin^2\left(\frac{\pi u}{2} \right)\, du = \int_0^p P_{1/2}^{-1}(u) \, du ,\quad \forall p \in [0,1],
\]
which can be verified by solving the integrals. Therefore, we have
\[
P_1 \leq_\mathrm{cx} P_{1/2}.
\]
On the other hand, using the differential entropy, we obtain $H(P_1) = \log(1) = 0$ and $H(P_{1/2}) = \log \left(\frac{\pi}{4} \right) \approx - 0.24 <0$.
Therefore, we have $P_1 \leq_\mathrm{cx} P_{1/2}$, but $\au(\delta_1) = 0 > -0.24 = \au(\delta_{1/2})$, which violates A3.
\end{proof}

\section{Analysis of Uncertainty Representation Methods}
\label{app:uncertainty_representation}
In this section, we analyze the already introduced uncertainty representation methods, i.e., deep ensembles and deep evidential regression, in more detail and provide an analysis for additional variants.

\subsection{Gaussian Deep Ensemble}
(Gaussian) deep ensembles, as popularized by \cite{NIPS2017_9ef2ed4b} provide a way to improve the performance of a neural network by simply training an ensemble of networks in a stochastic manner, therefore obtaining a more robust solution. When combined with a first-order predictive distribution, this leads to possible assessment of aleatoric- as well as epistemic uncertainty. Deep ensembles have been a gold standard for the past years \citep{ovadia2019trustmodelsuncertaintyevaluating, GANAIE2022105151} and have been applied to numerous tasks, such as computer vision \citep{Thuy_2024, ovadia2019trustmodelsuncertaintyevaluating}, image segmentation \citep{BUDDENKOTTE2023107096, MEHRTENS2023102914, HUETDASTARAC2024110545} or weather prediction \citep{NeuralNetworksforPostprocessingEnsembleWeatherForecasts, schulz2024aggregatingdistributionforecastsdeep}.

In the (univariate) deep ensemble setting \citep{NIPS2017_9ef2ed4b}, we have $Y\sim P_{\vvtheta} = \mathcal{N}(\mu, \sigma^2)$, with $\vvtheta\sim Q$. Here, $M$ different networks are trained in a stochastic manner, leading to a set of predictions $\vtheta_m = (\mu_m, \sigma_m^2), \ m=1, \ldots, M$. The ensemble acts as an empirical distribution $Q = \frac{1}{M} \sum_{m=1}^M \delta_{\vtheta_m}$, e.g., a mixture of (equally weighted) Dirac measures. For the predictive mixture, we obtain the Gaussian mixture 
\[
\overline{P} =  \frac{1}{M} \sum_{m=1}^M \mathcal{N}(\cdot\,  ; \mu_m, \sigma_m^2).
\]

\emph{Variance-based measures:}
For the variance-based measures, we obtain
\begin{align*}
\au(Q) &= \bE_Q[\bV(Y \mid \vvtheta)] =  \frac{1}{M} \sum_{m=1}^M \sigma_m^2, \\
\eu(Q) &= \bV_Q(\bE[Y\mid \vvtheta]) =  \frac{1}{M} \sum_{m=1}^M \mu_m^2 -\left(\frac{1}{M} \sum_{m=1}^M \mu_m \right)^2, \\
\tu(Q) &= \bV(Y) =  \frac{1}{M} \sum_{m=1}^M \left( \sigma_m^2+ \mu_m^2 \right) - \left(\frac{1}{M} \sum_{m=1}^M \mu_m \right)^2.
\end{align*}

\emph{Entropy-based measures:}
For the entropy-based measures, by the definition of differential entropy and the KL-divergence, we obtain
\begin{align*}
\au(Q) &= \bE_Q[H(Y \mid P_{\vvtheta})]  = \frac{1}{2M} \sum_{m=1}^M \log(2\pi e \sigma_m^2), \\
\eu(Q) &= \bE_Q[D_\mathrm{KL}(P_{\vvtheta}\| \overline{P})] =\frac{1}{M} \sum_{m=1}^M \int_{-\infty}^\infty p(t \mid \vtheta_m) \log \frac{p(t \mid \vtheta_m) }{\frac{1}{M}\sum_{l=1}^M p(t \mid \vtheta_l) }\, dt, \\
\tu(Q) &= H(\overline{P}) = \au(Q) + \eu(Q),
\end{align*}
where $p(x \mid \vtheta_m) = \mathcal{N}(x; \mu_m, \sigma_m^2)$. Here, $\eu(Q)$ (and therefore $\tu(Q)$) does not admit a closed-form solution, but can be approximated numerically.

\begin{figure}[ht]
\begin{subfigure}{0.5\linewidth}
\centering
\caption{$\mu_1 = 0, \ \mu_2 = 0$}
\includegraphics[width = \linewidth]{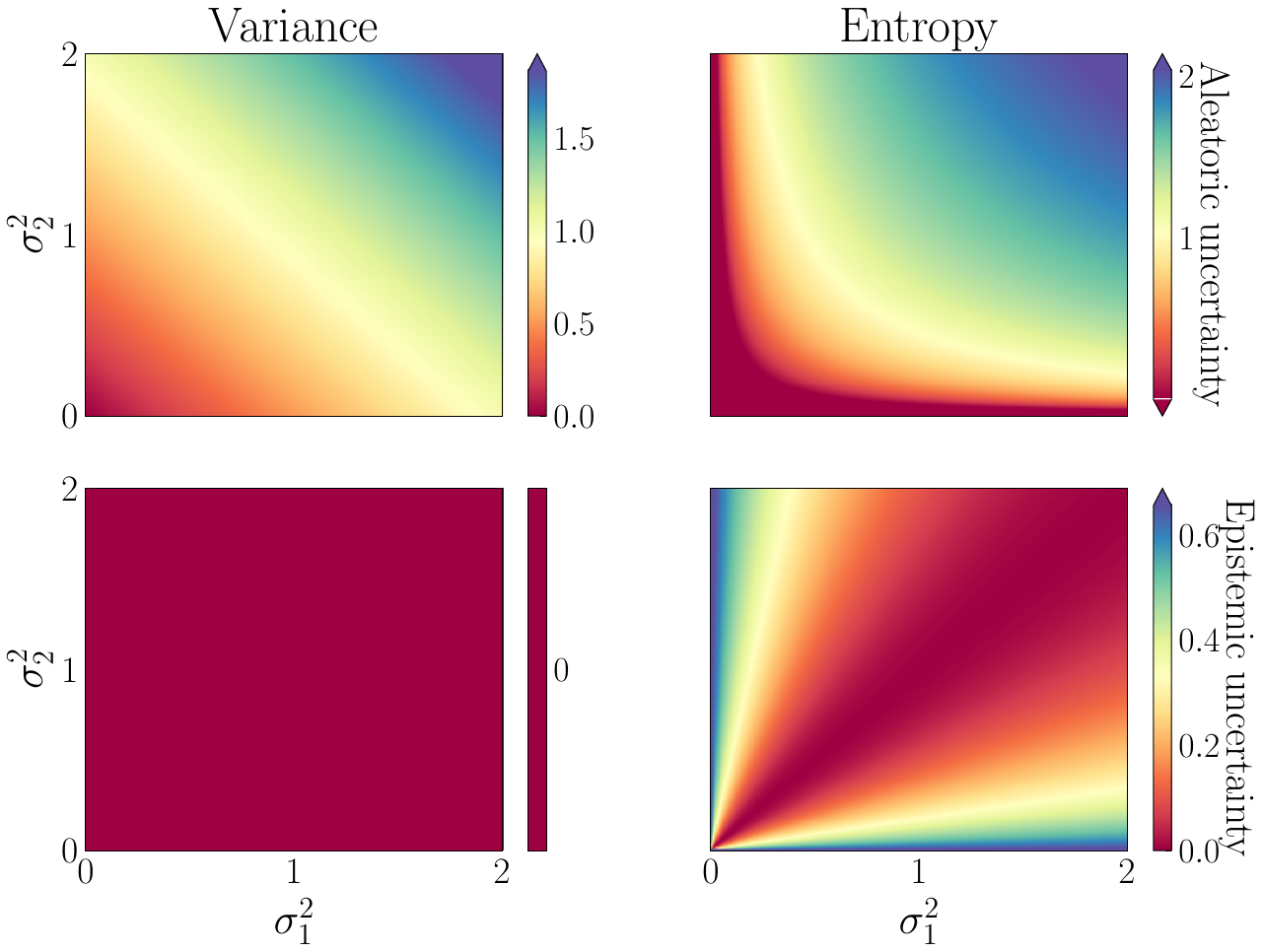}
\end{subfigure}
\begin{subfigure}{0.5\linewidth}
\centering
\caption{$\mu_1 = 2, \ \mu_2 = -2$}
\includegraphics[width = \linewidth]{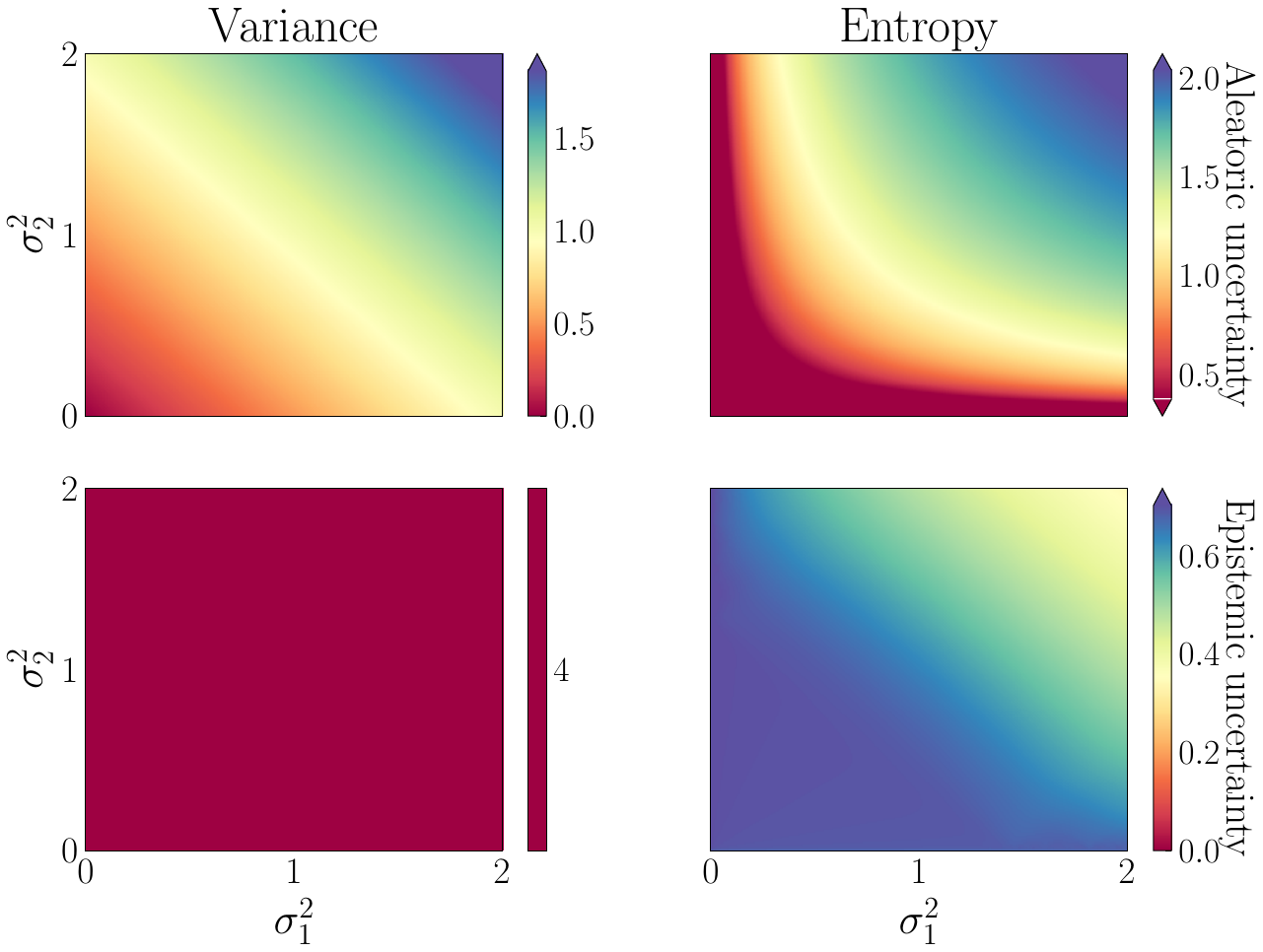}
\end{subfigure}
\begin{subfigure}{0.5\linewidth}
\centering
\caption{$\mu_1 = 0, \ \mu_2 = -2$}
\includegraphics[width = \linewidth]{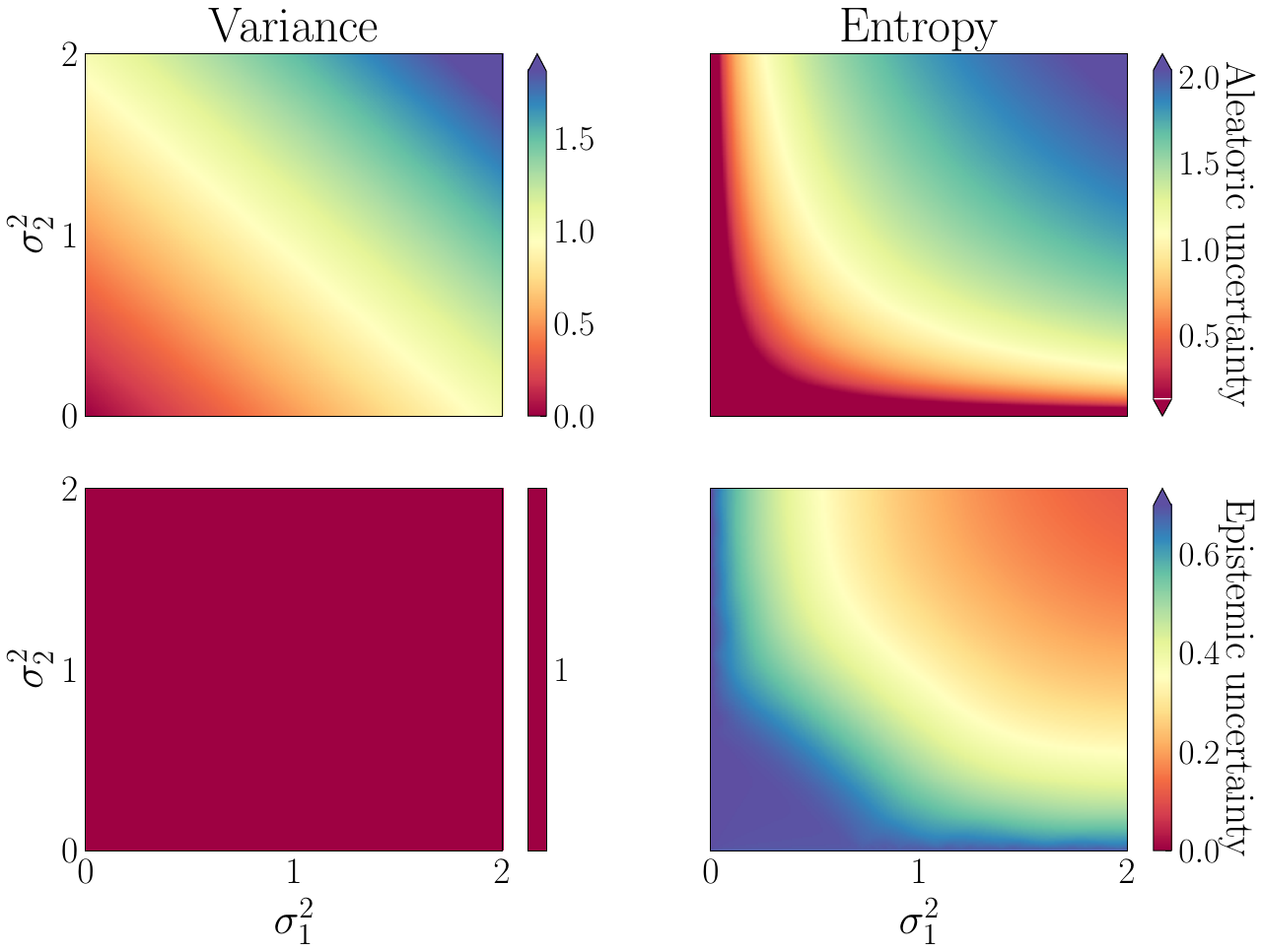}
\end{subfigure}
\begin{subfigure}{0.5\linewidth}
\centering
\caption{$\mu_1 = 2, \ \mu_2 = 0$}
\includegraphics[width = \linewidth]{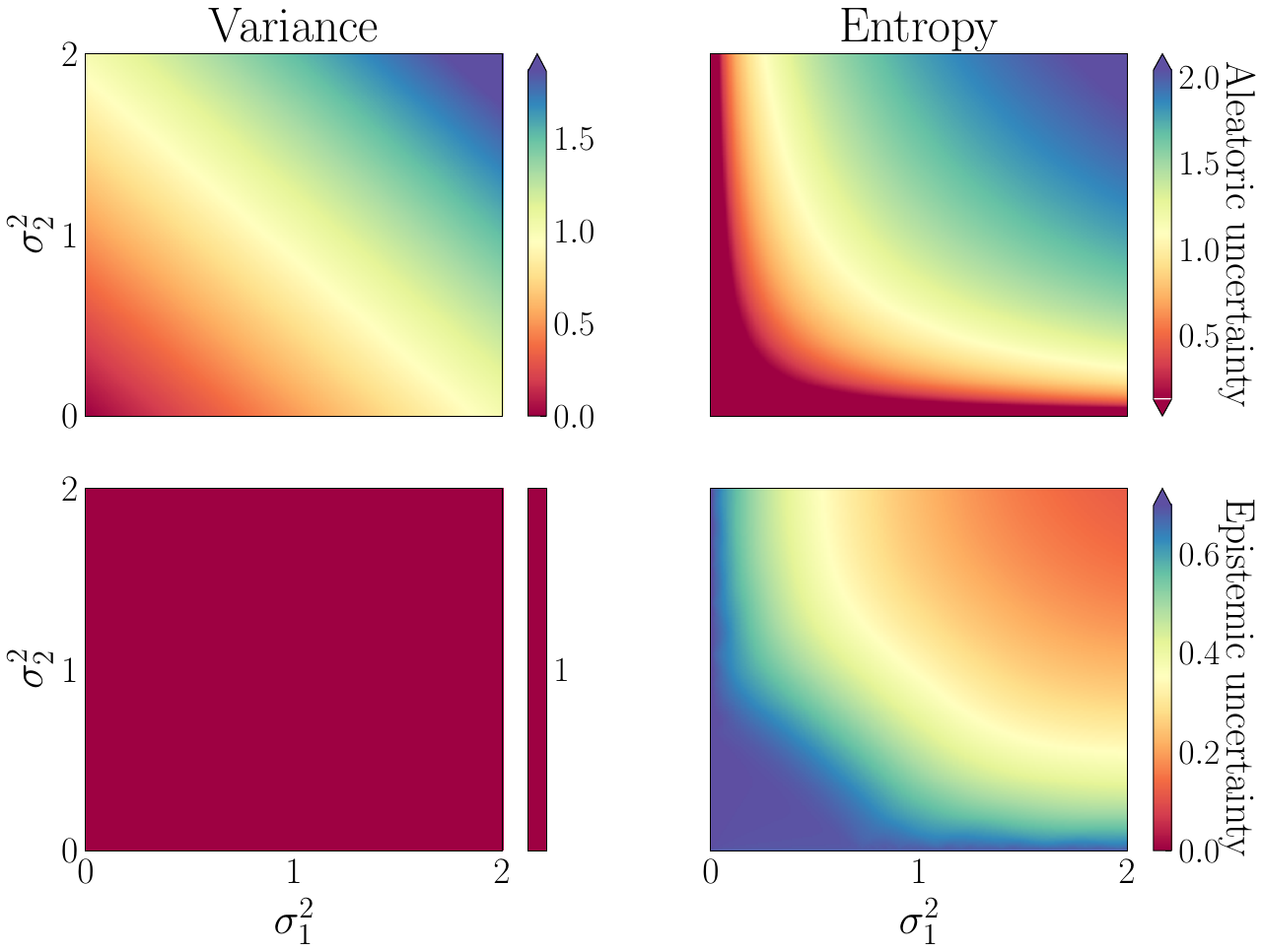}
\end{subfigure}
\caption{The figure shows the variance- and entropy-based different measures of uncertainty for a deep Gaussian ensemble $(\mu_m,\sigma_m^2)_{m=1}^2$ with two members across varying parameters.}
\label{fig:ensemble_2d_plot}
\end{figure}
\autoref{fig:ensemble_2d_plot} shows the closed-form solutions of the variance- and entropy-based measures (EU and AU) for a two-member Gaussian deep ensembles. It is visible that both measures differ significantly. Most notable, the variance-based measure for EU does not change across $\sigma_1^2, \sigma_2^2$, as it only measures uncertainty based on the first-order moments, which is a direct implication of Proposition~\ref{prop:var_a1}.
In addition, both measures provide very different estimates for AU. For the variance-based measure the estimate depends, by definition, on $\sigma_1^2, \sigma_2^2$ in a linear way, while for the entropy-based measure the dependence is in a significantly different, nonlinear way, as visualized in the first row of each subplot in \autoref{fig:ensemble_2d_plot}.

\subsection{Laplacian Deep Ensemble}
While research and applications have mainly focused on the Gaussian case \citep{GANAIE2022105151}, for the deep ensemble one can essentially consider any learnable parametric first-order distribution. The variance- and entropy-based uncertainty measures transfer in the same way, although closed-form solutions might not necessarily be available. As an example, we analyze how the different measures behave for a deep ensemble with a first-order Laplace distribution, which was used by \cite{KIANISHAHVANDI2025105818} to model and predict deviation of time differences between earth's rotation time and coordinated universal time.

The setting is the same as the Gaussian deep ensemble setting, but with a predictive Laplace distribution, i.e., $Y\sim P_{\vvtheta} = \mathrm{Laplace}(\mu, \eta), \vvtheta\sim Q$ with density $$p(y\mid \vtheta) = \frac{1}{2\eta} \exp\left( - \frac{|y - \mu|}{\eta} \right), \ {\vtheta = (\mu, \eta)^\top \in \Ree \times \Ree_{>0}}.$$
For the Laplace distribution it holds that $\bE[\mathrm{Laplace}(\mu, \eta)] = \mu$ and $\bV(\mathrm{Laplace}(\mu, \eta)) = 2\eta^2$. For the predictive mixture, we obtain
\[
\overline{P} =  \frac{1}{M} \sum_{m=1}^M \mathrm{Laplace}(\cdot \, ; \mu_m, \eta_m).
\]

\emph{Variance-based measures:}
For the variance-based measures, we obtain
\begin{align*}
\au(Q) &= \bE_Q[\bV(Y \mid \vvtheta)] =  \frac{1}{M} \sum_{m=1}^M 2\eta_m^2, \\
\eu(Q) &= \bV_Q(\bE[Y\mid \vvtheta]) =  \frac{1}{M} \sum_{m=1}^M \mu_m^2 -\left(\frac{1}{M} \sum_{m=1}^M \mu_m \right)^2, \\
\tu(Q) &= \bV(Y) =  \frac{1}{M} \sum_{m=1}^M \left( 2\eta_m^2+ \mu_m^2 \right) - \left(\frac{1}{M} \sum_{m=1}^M \mu_m \right)^2.
\end{align*}

\emph{Entropy-based measures:}
For the entropy-based measures, by the definition of differential entropy and the KL-divergence, we obtain
\begin{align*}
\au(Q) &= \bE_Q[H(Y \mid P_{\vvtheta})]  = 1+\frac{1}{M} \sum_{m=1}^M \log(2\eta_m) , \\
\eu(Q) &= \bE_Q[D_\mathrm{KL}(P_{\vvtheta}\| \overline{P})] =\frac{1}{M} \sum_{m=1}^M \int_{-\infty}^\infty p(t \mid \vtheta_m) \log \frac{p(t \mid \vtheta_m) }{\frac{1}{M}\sum_{l=1}^M p(t \mid \vtheta_l) }\, dt, \\
\tu(Q) &= H(\overline{P}) = \au(Q) + \eu(Q).
\end{align*}
Here, $\eu(Q)$ (and therefore $\tu(Q)$) does not admit a closed-form solution, but can be approximated numerically.

\begin{figure}[ht]
\begin{subfigure}{0.5\linewidth}
\centering
\caption{$\mu_1 = 0, \ \mu_2 = 0$}
\includegraphics[width = \linewidth]{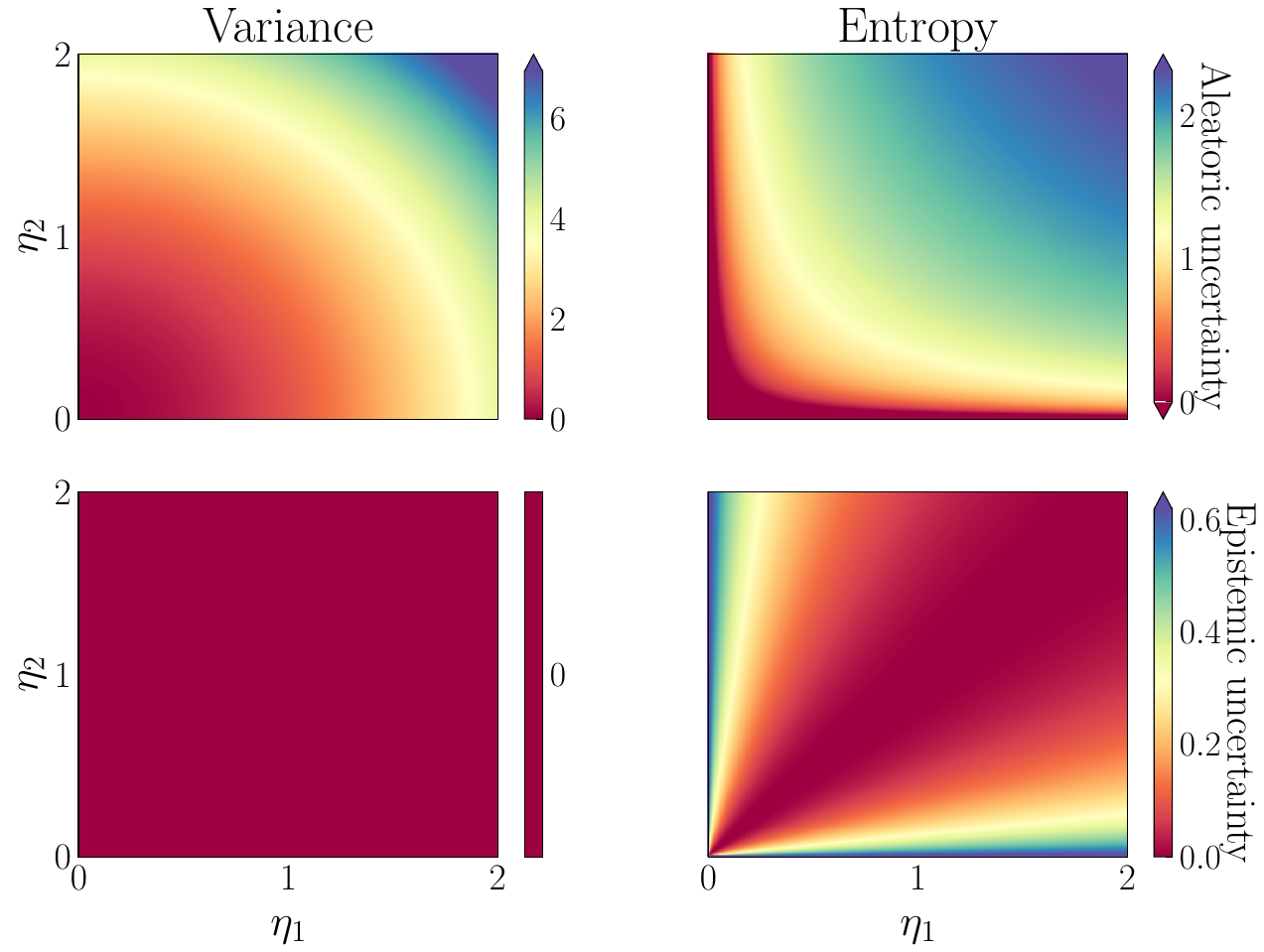}
\end{subfigure}
\begin{subfigure}{0.5\linewidth}
\centering
\caption{$\mu_1 = 2, \ \mu_2 = -2$}
\includegraphics[width = \linewidth]{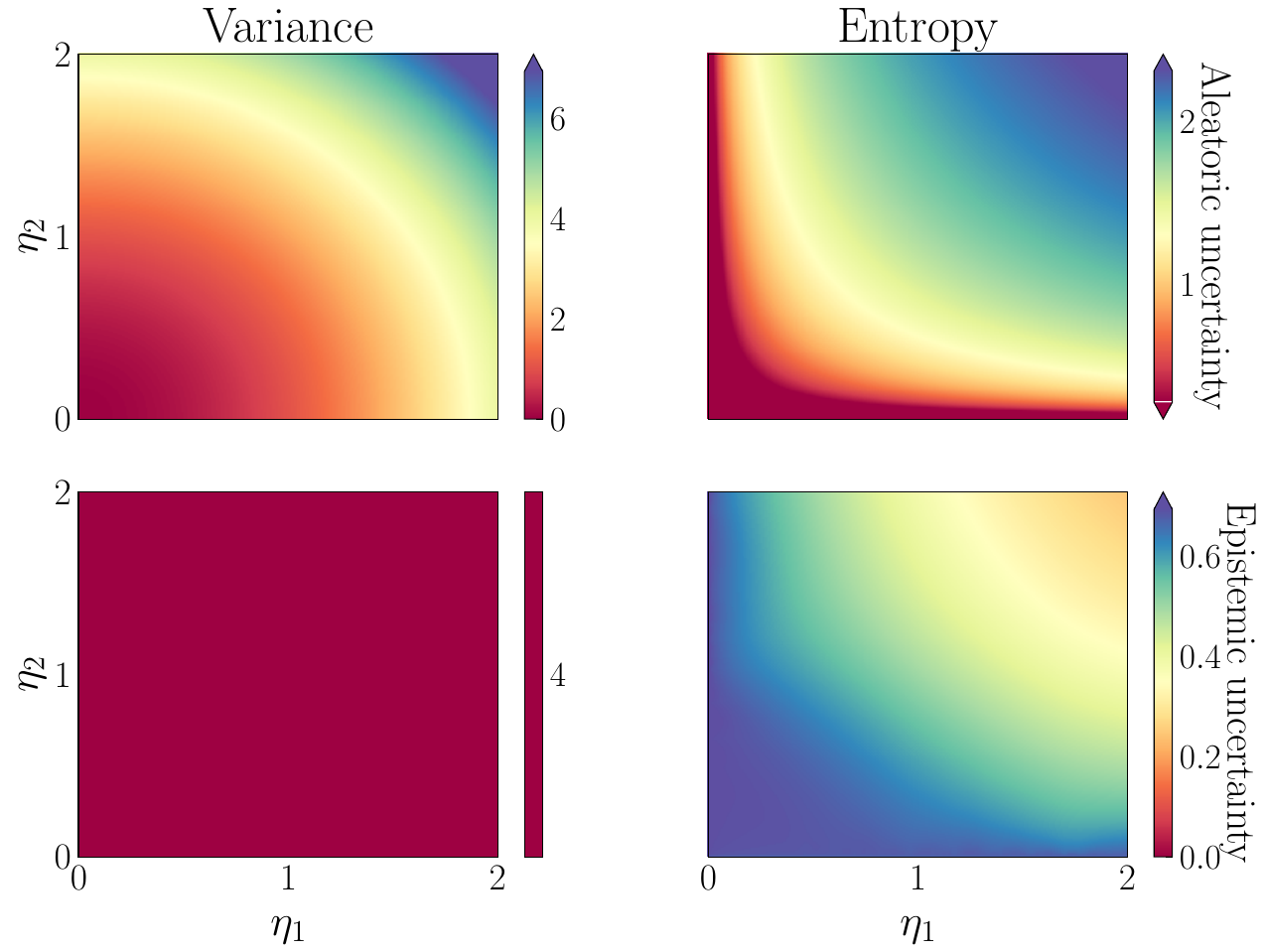}
\end{subfigure}
\begin{subfigure}{0.5\linewidth}
\centering
\caption{$\mu_1 = 0, \ \mu_2 = -2$}
\includegraphics[width = \linewidth]{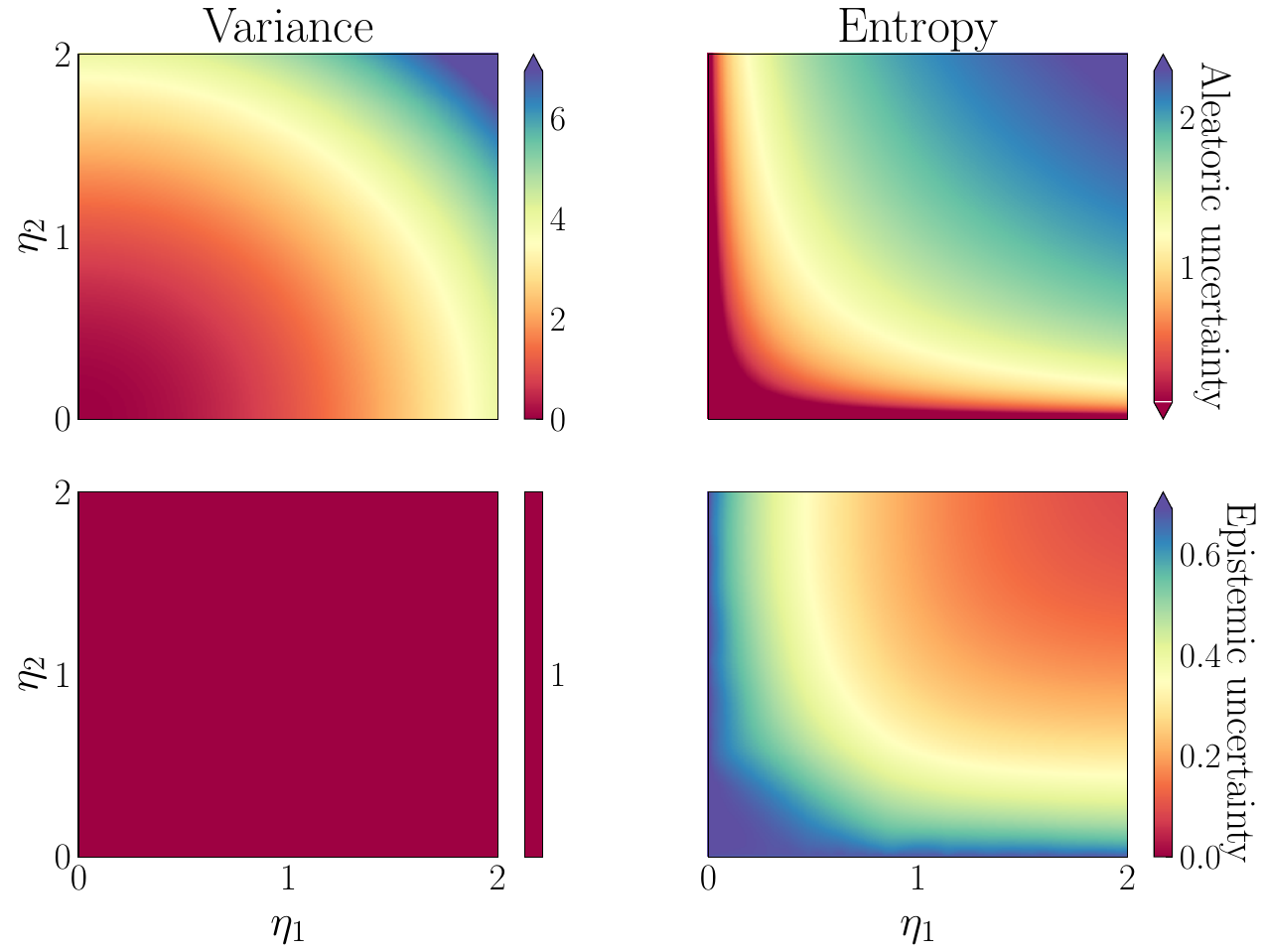}
\end{subfigure}
\begin{subfigure}{0.5\linewidth}
\centering
\caption{$\mu_1 = 2, \ \mu_2 = 0$}
\includegraphics[width = \linewidth]{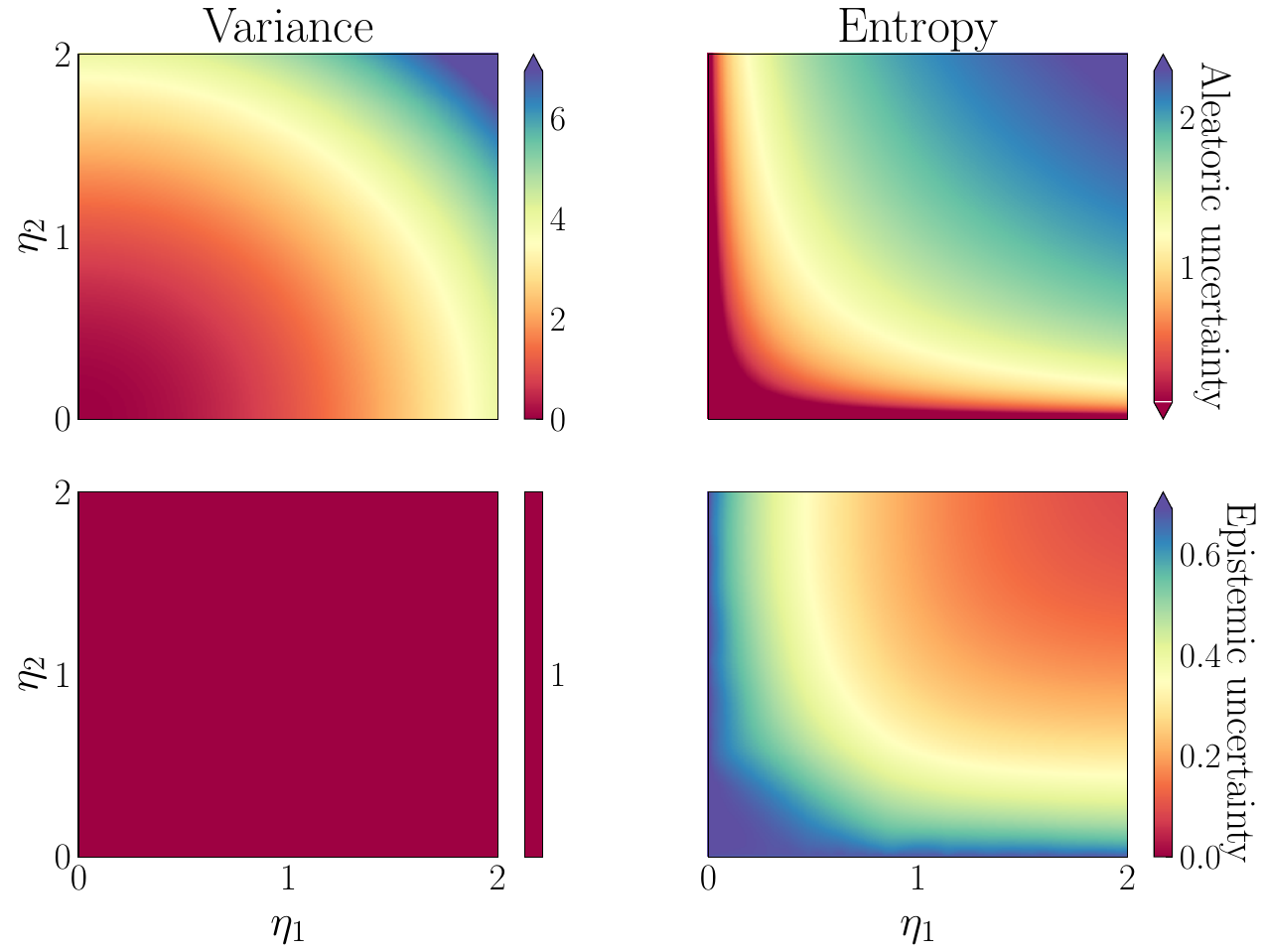}
\end{subfigure}
\caption{The figure shows the variance- and entropy-based different measures of uncertainty for a deep Laplacian ensemble $(\mu_m,\sigma_m^2)_{m=1}^2$ with two members across varying parameters.}
\label{fig:ensemble_laplace_2d_plot}
\end{figure}
\autoref{fig:ensemble_laplace_2d_plot} shows the closed-form solutions of the variance- and entropy-based measures (EU and AU) for a two-member Laplacian deep ensembles. It is evident that both measures differ heavily. Similar to the Gaussian case, the variance-based measure of EU does not change across $\eta_1, \eta_2$.
However, even for AU, both measures differ significantly. For both, AU depends on $\eta_1, \eta_2$ in a nonlinear way, however the corresponding curvatures are different, as visible in the two first rows of each plot.

\subsection{Univariate Deep Evidential Regression}
Deep evidential regression, introduced by \cite{amini2020deep} is a widely used method to provide uncertainty estimates and is based on the conjugate prior of a normal distribution.
Recall that in this setting, we have $P_{\vvtheta} = \mathcal{N}(\mu, \sigma^2)$ with $\vvtheta\sim Q$ and $q(\mu, \sigma^2) = q(\mu) \, q(\sigma^2) = \mathcal{N}(\mu; \gamma, \sigma^2 \upsilon^{-1}) \, \Gamma^{-1} (\sigma^2; \alpha, \beta)$. The predictive mixture is given via a Student's t distribution $$\overline{P} = t_{2\alpha}\left(\cdot \,; \gamma, \frac{\beta(1+\upsilon)}{\alpha \upsilon} \right).$$

\emph{Variance-based measures:}
\citep{amini2020deep} show that by calculating the corresponding moments, the variance-based uncertainty measures are given as
\begin{align*}
\au(Q) &= \bE_Q[\bV(Y \mid \vvtheta)] = \frac{\beta}{\alpha - 1}, \\
\eu(Q) &= \bV_Q(\bE[Y\mid \vvtheta]) = \frac{\beta}{\upsilon(\alpha - 1)}, \\
\tu(Q) &= \bV(Y) =  \frac{\beta}{\alpha - 1}(1+ \upsilon^{-1}).
\end{align*}

\emph{Entropy-based measures:}
For the entropy-based measures, we obtain
\begin{align*}
\au(Q) &=  \frac{1}{2} (\log(2\pi e)+\log(\beta)- \psi(\alpha)) , \\
\eu(Q) &= \frac{2\alpha+1}{2} \psi\left(\alpha + \frac{1}{2}\right) - \alpha \psi \left(\alpha \right)  - \frac{1}{2} \log(\pi e) + \log B\left(\alpha, \frac{1}{2} \right) + \frac{1}{2} \log \left(\frac{1+\upsilon}{\upsilon} \right), \\
\tu(Q) &= \frac{2\alpha+1}{2}\left( \psi\left(\alpha + \frac{1}{2}\right) - \psi \left(\alpha \right) \right) + \log B\left(\alpha, \frac{1}{2} \right)+ \frac{1}{2}\log \left( \frac{2\beta(1+ \upsilon)}{ \upsilon}\right).
\end{align*}
This can be derived by the properties of the Normal-Inverse-Gamma distribution. For aleatoric uncertainty, we have
\begin{align*}
    \mathrm{AU}(Q) &= \mathbb{E}_{q(\vvtheta)}[H(P_{\vvtheta})] = \mathbb{E}_{q(\vvtheta)}\left[\frac{1}{2} \log(2\pi e\sigma^2) \right] = \mathbb{E}_{\sigma^2\sim \Gamma^{-1}(\alpha, \beta)}\left[\frac{1}{2} \log(2\pi e \sigma^2) \right] \\
    &= \frac{1}{2} \log(2\pi e) + \frac{1}{2}\mathbb{E}_{\sigma^2\sim \Gamma^{-1}(\alpha, \beta)}\left[\log(\sigma^2) \right] = \frac{1}{2} (\log(2\pi e)+\log(\beta)- \psi(\alpha)),
\end{align*}
where we used that for $S \sim \Gamma^{-1}(\alpha,\beta)$ we have $\mathbb{E}[\log(S)] = \log(\beta) - \psi(\alpha)$, where $\psi$ denotes the digamma function. For total uncertainty, we use
\begin{align*}
\tu(Q) &= H(\overline{P}) = H \left(t_{2\alpha}\left(\gamma, \frac{\beta (1+ \upsilon)}{\alpha \upsilon}\right)\right)\\
&= \frac{2\alpha+1}{2}\left( \psi\left(\alpha + \frac{1}{2}\right) - \psi \left(\alpha \right) \right)  + \log B\left(\alpha, \frac{1}{2} \right)+ \frac{1}{2}\log \left( \frac{2\beta(1+ \upsilon)}{ \upsilon}\right),
\end{align*}
where $B(\cdot, \cdot)$ is the Beta function. Then epistemic uncertainty can be obtained via ${\eu(Q) = \tu(Q) - \au(Q)}$.
\begin{figure}[ht]
\begin{subfigure}{0.5\linewidth}
\centering
\caption{$\nu = 0.1$}
\includegraphics[width = \linewidth]{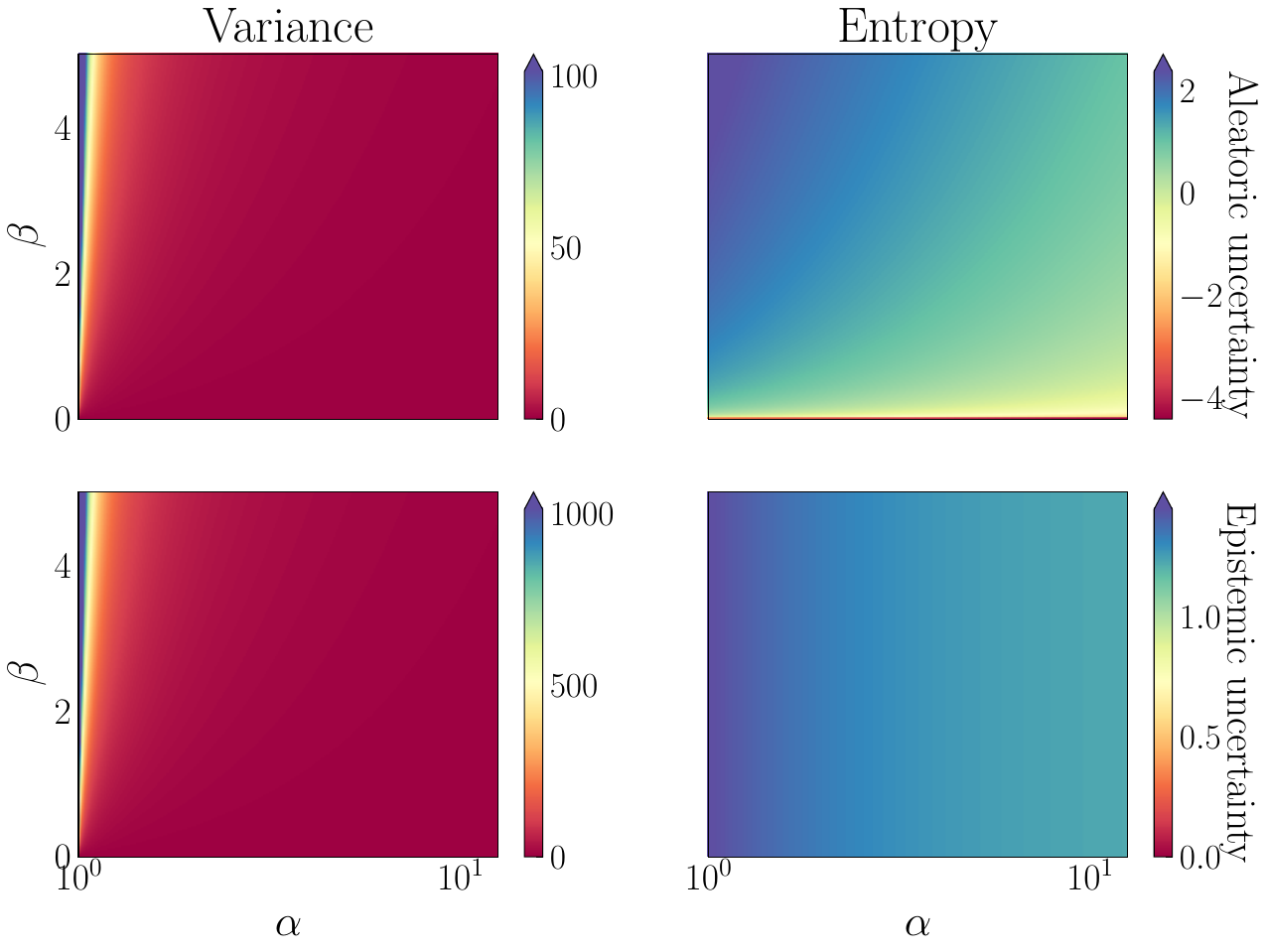}
\end{subfigure}
\begin{subfigure}{0.5\linewidth}
\centering
\caption{$\nu =  0.5$}
\includegraphics[width = \linewidth]{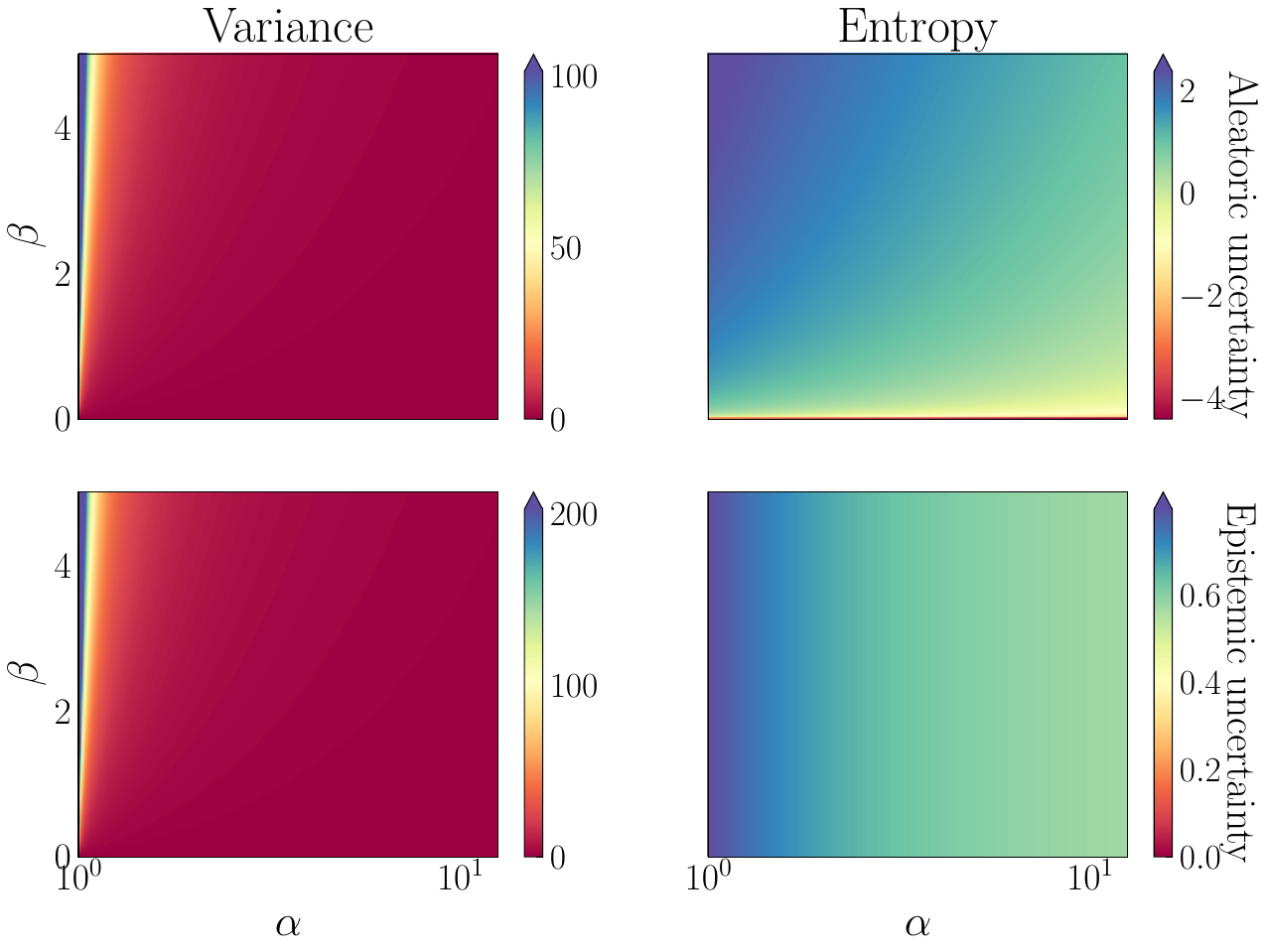}
\end{subfigure}
\begin{subfigure}{0.5\linewidth}
\centering
\caption{$\nu = 2$}
\includegraphics[width = \linewidth]{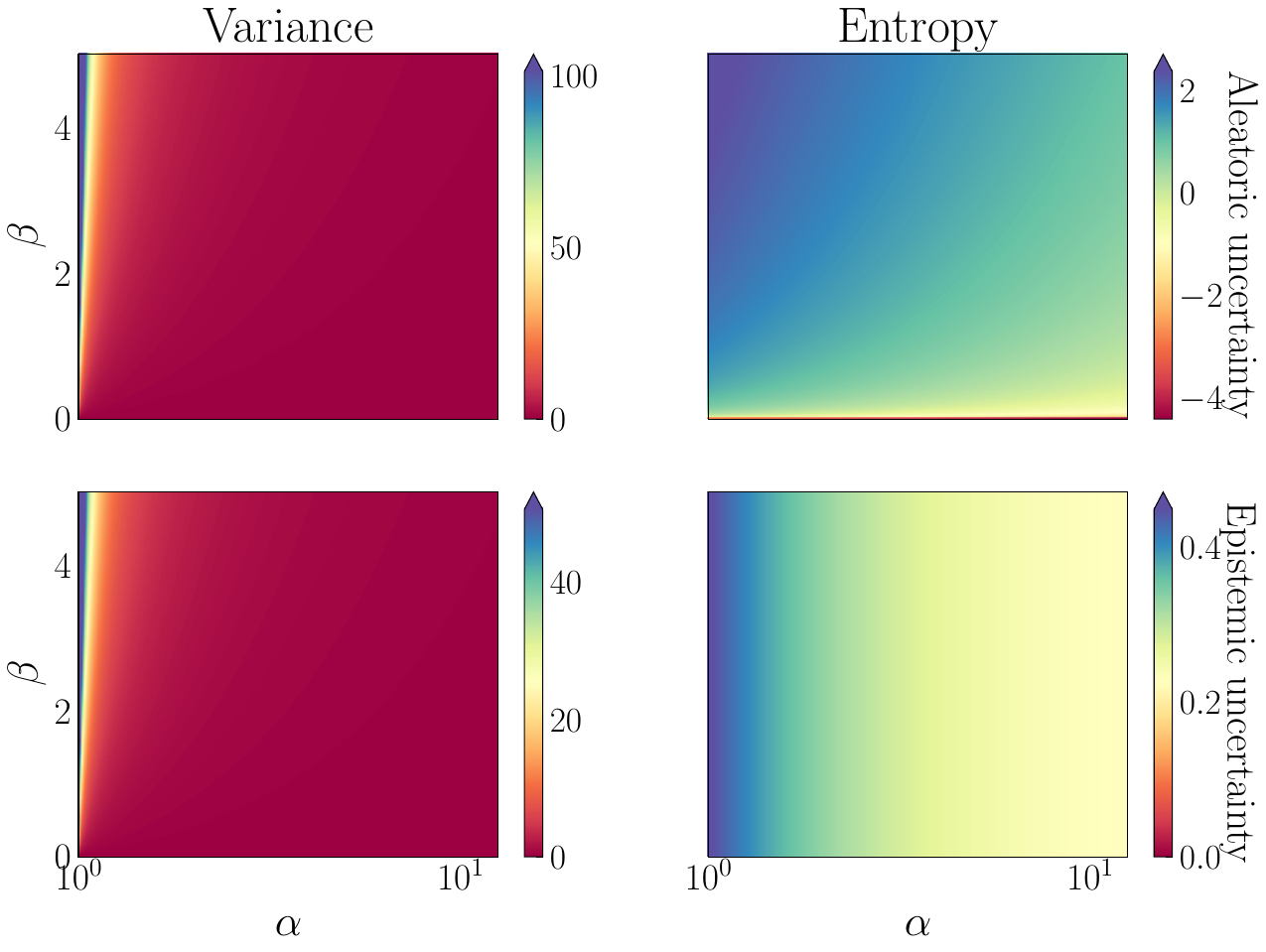}
\end{subfigure}
\begin{subfigure}{0.5\linewidth}
\centering
\caption{$\nu = 5$}
\includegraphics[width = \linewidth]{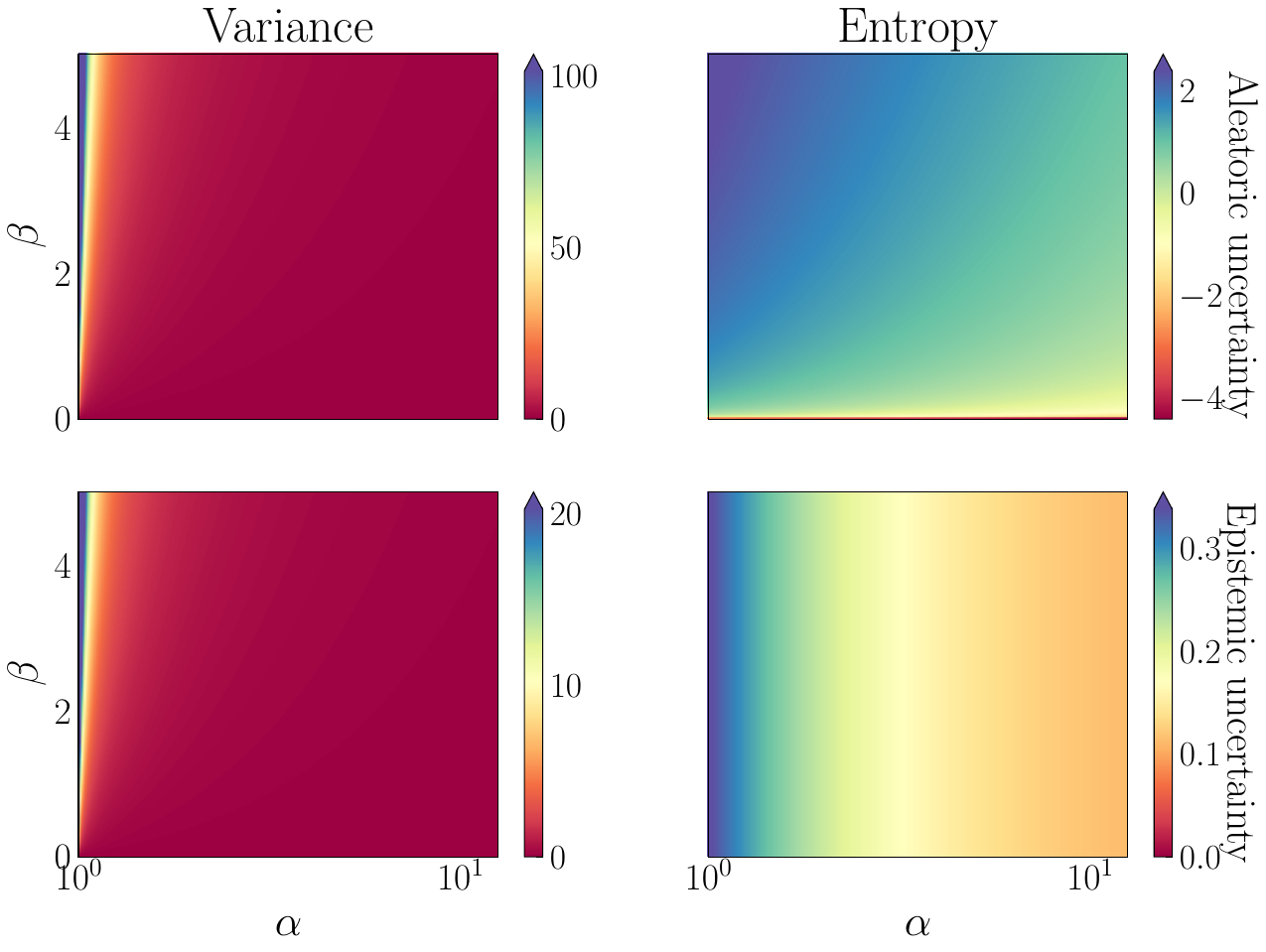}
\end{subfigure}
\caption{The figure shows the variance- and entropy-based measures of uncertainty for the deep evidential regression setting, across varying parameters.}
\label{fig:der_2d_plot}
\end{figure}
\autoref{fig:der_2d_plot} shows the closed-form solutions of the variance- and entropy-based measures (EU and AU) for the deep evidential regression setting.
It is clear that both measures behave very differently. In particular, for the variance-based measures EU and AU diverge for $\alpha \downarrow 1$, while the entropy-based measures are well defined for all $\alpha >0$.
Furthermore, the entropy-based EU estimate does not depend on $\beta$, which corresponds to the scale parameter of the first-order noise.

\subsection{Multivariate Deep Evidential Regression}
\cite{meinert2022multivariatedeepevidentialregression} extend the deep evidential regression to the multivariate setting, by considering the corresponding conjugate prior of a multivariate normal, which is given by a Normal-Inverse-Wishart distribution.
In this setting, we have $\boldsymbol{Y} \in \Ree^d$ and $\boldsymbol{Y} \sim P_{\vvtheta} = \mathcal{N}(\boldsymbol{\mu}, \boldsymbol\Sigma)$ with $\vvtheta\sim Q$. The corresponding densities are given as $q(\boldsymbol{\mu}, \boldsymbol{\Sigma}) = q(\boldsymbol{\mu}) q(\boldsymbol{\Sigma})$ with
\[
q(\boldsymbol{\mu}) = \mathcal{N}(\boldsymbol{\mu}; \boldsymbol{\mu}_0, \boldsymbol{\Sigma}/\kappa), \qquad q(\boldsymbol{\Sigma}) = \mathcal{W}^{-1}(\boldsymbol{\Psi},\nu),
\]
where $\mathcal{W}^{-1}$ denotes the inverse-Wishart distribution and we have $\kappa>0, \nu > d+1, \boldsymbol{\Psi} \in \Ree^{d \times d}$. The predictive mixture is given via a multivariate $t$-distribution with $\nu - d+1$ degrees of freedom, i.e., 
\[
\overline{P} = t_{\nu - d+1}\left(\cdot \,; \boldsymbol{\mu}_0, \frac{1+\kappa}{\kappa(\nu - d+1)} \boldsymbol{\Psi} \right).
\]

\emph{Variance-based measures:}
Similar to the univariate case, the variance-based uncertainty measures can be calculated from the moments of the underlying (Normal-Inverse-Wishart) distribution as
\begin{align*}
\au(Q) &= \bE_Q[\bV(\boldsymbol Y \mid \vvtheta)] = \bE_Q[\tr(\mathrm{Cov(\boldsymbol{Y}\mid \vvtheta)})] = \frac{\tr(\boldsymbol\Psi)}{\nu -d-1}, \\
\eu(Q) &= \bV_Q(\bE[\boldsymbol Y\mid \vvtheta]) = \tr(\mathrm{Cov}(\bE[\boldsymbol{Y}\mid \vvtheta])) = \frac{\tr(\boldsymbol\Psi)}{\kappa(\nu -d-1)}, \\
\tu(Q) &= \bV(\boldsymbol Y) = \frac{\tr(\boldsymbol\Psi)}{\nu -d-1} \left( 1+ \frac{1}{\kappa} \right).
\end{align*}

\emph{Entropy-based measures:}
For the entropy-based measures, we obtain
\begin{align*}
\au(Q) &= \frac{d}{2} \log(\pi e) + \frac{1}{2} \log (|\boldsymbol{\Psi}|) - \frac{1}{2} \psi_d\left(\frac{\nu}{2} \right) , \\
\eu(Q)&= \frac{d}{2}\log\left(\tfrac{1+1/\kappa}{e}\right)
 - \log\Gamma\left(\frac{\nu+1}{2}\right)
 + \log\Gamma\left(\frac{\nu-d+1}{2}\right) \\
&\quad
 + \frac{\nu+1}{2}\left[
   \psi\left(\frac{\nu+1}{2}\right)
   - \psi\left(\frac{\nu-d+1}{2}\right)
 \right]
 + \frac{1}{2}\psi_d\left(\frac{\nu}{2}\right),\\
\tu(Q) &= \frac{1}{2}\log|\Psi|
 + \frac{d}{2}\log\left(\pi\left(1+\frac{1}{\kappa}\right)\right)
 -\log\Gamma\left(\frac{\nu+1}{2}\right)
 + \log\Gamma\left(\frac{\nu-d+1}{2}\right) \\
&\quad
 + \frac{\nu+1}{2}\left[
   \psi\left(\frac{\nu+1}{2}\right)
   - \psi\left(\frac{\nu-d+1}{2}\right)
 \right],
\end{align*}
where $\Gamma(\cdot)$ denotes the Gamma function, $\psi(\cdot)$ the digamma function, and $\psi_d(\cdot)$ the multivariate digamma function.
The derivations are similar to the univariate case and based on the differential entropy of a multivariate $t$-distribution.

\end{document}